%% file: main.tex
\documentclass{article}
\usepackage{amsmath}
\usepackage{amssymb}
\usepackage{xcolor}
\usepackage{booktabs}
\usepackage{multirow}
\usepackage{graphicx}
\usepackage{adjustbox}
\usepackage{wrapfig}
\usepackage{algorithm}
\usepackage{algpseudocode}
\usepackage{caption}
\newcommand{\methodname}{SC3-Eval}
\newif\ifshowedits
\showeditsfalse
\newcommand{\edit}[1]{\ifshowedits\textcolor{teal}{#1}\else#1\fi}

\usepackage[preprint]{corl_2026} % Uncomment for pre-prints (e.g., arxiv); This is like ``final'', but will remove the CORL footnote.

\makeatletter
\if@preprinttype
  \newcommand{\projecturl}{\url{https://weichengtseng.github.io/sc3-eval/}}
\else
  \if@conferencefinal
    \newcommand{\projecturl}{\url{https://weichengtseng.github.io/sc3-eval/}}
  \else
    \newcommand{\projecturl}{\url{https://miscsubmission.github.io/sc3-eval/}}
  \fi
\fi
\makeatother

\title{SC3-Eval: Evaluating Robot Foundation Models via Self-Consistent Video Generation}
\author{
  Wei-Cheng Tseng$^{1,2,3}$ \\
  \And
  Gashon Hussein$^{4}$ \\
  \And
  Yuzhu Dong$^{3}$ \\
  \And
  Allen Z. Ren$^{4}$ \\
  \And
  Lucy X. Shi$^{4,5}$ \\
  \And
  XuDong Wang$^{4}$ \\
  \And
  Sergey Levine$^{4,6}$ \\
  \And
  Zhaoshuo Li$^{3}$ \\
  \And
  Jinwei Gu$^{3}$ \\
  \And
  Florian Shkurti$^{1,2}$ \\
  \And
  Ming-Yu Liu$^{3}$ \\
  \And
  Quan Vuong$^{4}$ \\
  \AND
  \normalfont $^{1}$University of Toronto \quad
  $^{2}$Vector Institute \quad
  $^{3}$NVIDIA \quad
  $^{4}$Physical Intelligence \\
  $^{5}$Stanford University \quad
  $^{6}$UC Berkeley
}
\begin{document}
\maketitle

%===============================================================================

\input{sources/abstract}


%===============================================================================

\input{sources/sec1_intro}
\input{sources/sec4_related_works}
\input{sources/sec2_method}

\input{sources/sec3_experiments}
\input{sources/sec5_limitations}
\input{sources/sec6_conclusions}
% \input{sources/todo}

%===============================================================================

\clearpage
% The acknowledgments are automatically included only in the final and preprint versions of the paper.
% \acknowledgments{We would like to appreciate .}

%===============================================================================

\bibliography{example}  % .bib

\newpage
\appendix
\input{sources/appendix}

\end{document}

%% file: sources/abstract.tex
\begin{abstract}
Evaluating generalist robot manipulation policies in the real world is expensive, slow, and difficult to scale. Action-conditioned video world models offer a scalable alternative by simulating policy rollouts. Autoregressive rollouts accumulate compounding errors, observations across multiple camera views must remain mutually consistent, and the evaluator must generalize to policies whose behaviors lie outside the training distribution. We address these challenges with \methodname, a \emph{self-consistent video generation} recipe that adapts a pre-trained video foundation model into an accurate policy evaluator by enforcing three complementary forms of consistency. First, \emph{forward-inverse dynamics consistency} jointly trains the model to predict frames from actions and to recover actions from frames, anchoring generated rollouts to a physically plausible action manifold and counteracting the drift a forward-only model cannot penalize. Second, \emph{cross-view consistency} trains the model to inpaint each camera view from the other, keeping the multi-camera observation coherent over long rollouts without any explicit memory mechanism. Third, \emph{test-time consistency} reuses the inverse dynamics mode at inference as a per-action-chunk uncertainty signal that terminates rollouts whose generated frames drift away from the requested actions. 
We also demonstrate \methodname{} rollouts reproduce the failure modes that policies exhibit in real-world rollouts, supporting fine-grained diagnostic comparison rather than aggregate ranking alone.
Across seven real-world vision-language-action policies, \edit{\methodname{} attains} a closed-loop Pearson correlation of $0.929$ and MMRV of $0.119$, outperforming three strong prior video-model-based baselines, and generalizes to new tasks. 
% and supplies context for automatic vision-language-model annotation. 
\vspace{-6pt}
\end{abstract}

%% file: sources/sec1_intro.tex
\vspace{-6pt}
\section{Introduction}
\vspace{-6pt}
Evaluating generalist robot manipulation policies in the real world is expensive, slow, and difficult to scale~\cite{yang2026robolab,polaris}. Each evaluation requires physical robot time, careful environment resets, and human supervision. These costs compound quickly when assessing generalist policies across diverse tasks and environments, motivating scalable alternatives that can reliably predict real-world policy performance without exhaustive physical rollouts.

Action-conditioned video world models have recently emerged as a promising substrate for this purpose~\cite{irasim, worldgym, tseng2025scalable, ctrlworld}. Conditioned on an initial multi-view observation and a sequence of actions, such a model simulates plausible future frames so that a policy can be rolled out and scored without touching a physical robot. However, applying this paradigm to real-world manipulation is difficult for three reasons. First, autoregressive rollouts accumulate drift as small errors compound over long horizons. Second, multi-camera observations must remain mutually consistent, or the policy may be misled by a visually incoherent scene. Third, the evaluator must generalize to policies whose behavior differs from its training distribution.

We address these challenges with \methodname, a \emph{self-consistent video generation} recipe that adapts a pre-trained video foundation model into a faithful closed-loop policy evaluator by enforcing three complementary forms of consistency (Fig.~\ref{fig:pipeline}). \emph{Forward-inverse dynamics consistency} jointly trains the model to predict frames from actions and to recover actions from frames, anchoring generated rollouts to a physically plausible action manifold and counteracting the drift a forward-only model lacks the signal to penalize. \emph{Cross-view consistency} trains the model to inpaint each \edit{held-out} camera view from the \edit{remaining one}, keeping the multi-camera observation coherent over long autoregressive rollouts without any explicit memory mechanism. \emph{Test-time consistency} reuses the inverse dynamics mode at inference as a per-action-chunk uncertainty signal that terminates rollouts whose generated frames drift away from the requested actions, and additionally supplies context for automatic vision-language-model annotation.

We instantiate this recipe on top of the pre-trained video backbone with a unified dynamics model~\cite{li2025uva,zhu2025uwm} and train it on a $381$-hour real-world table bussing manipulation dataset. To comprehensively evaluate the faithfulness of the policy evaluation pipeline, we focus on table bussing, which involves interactions between the robot and $12$ distinct objects spanning utensils and trash items. 
% Table bussing requires the robot to place utensils in the utensil bin and trash items in the trash can. Reverse table bussing keeps the same scene but swaps the destinations, sending utensils to the trash can and trash items to the utensil bin, forming a controlled out-of-distribution split that probes generalization beyond training task semantics rather than only beyond pixels or motions. 
In experiments on seven vision-language-action \edit{(VLA)} policy checkpoints, {\methodname} attains a closed-loop Pearson correlation of $0.929$ and MMRV of $0.119$, outperforming three strong existing video-model-based baselines including Ctrl-World~\citep{ctrlworld}, IRASim~\citep{irasim}, and Cosmos-Predict 2.5~\citep{cosmospredict25}. It generalizes to a held-out task variant absent from the training data, and beyond predicting aggregate success rates, it reproduces the specific failure modes that policies exhibit in real-world rollouts. Qualitative rollout videos are available on the project \edit{website}\footnote{\edit{\projecturl}}.

\input{sources/figures/fig_pipeline}
\textbf{Contributions.} First, we propose \methodname, a self-consistent video generation recipe whose three consistency objectives together adapt a pre-trained video foundation model into \edit{an} accurate closed-loop policy evaluator. Second, we introduce an uncertainty-driven early-termination criterion derived without modification to the training procedure from the inverse dynamics mode, which improves closed-loop reliability without additional supervision. Third, in real-world experiments on seven VLA policy checkpoints, {\methodname} achieves state-of-the-art policy evaluation performance, generalizes to an out-of-distribution task semantic, and reproduces per-trajectory failure modes rather than only aggregate success rates.

%% file: sources/figures/fig_pipeline.tex
\begin{figure}[t]
    \centering
    \includegraphics[width=\textwidth]{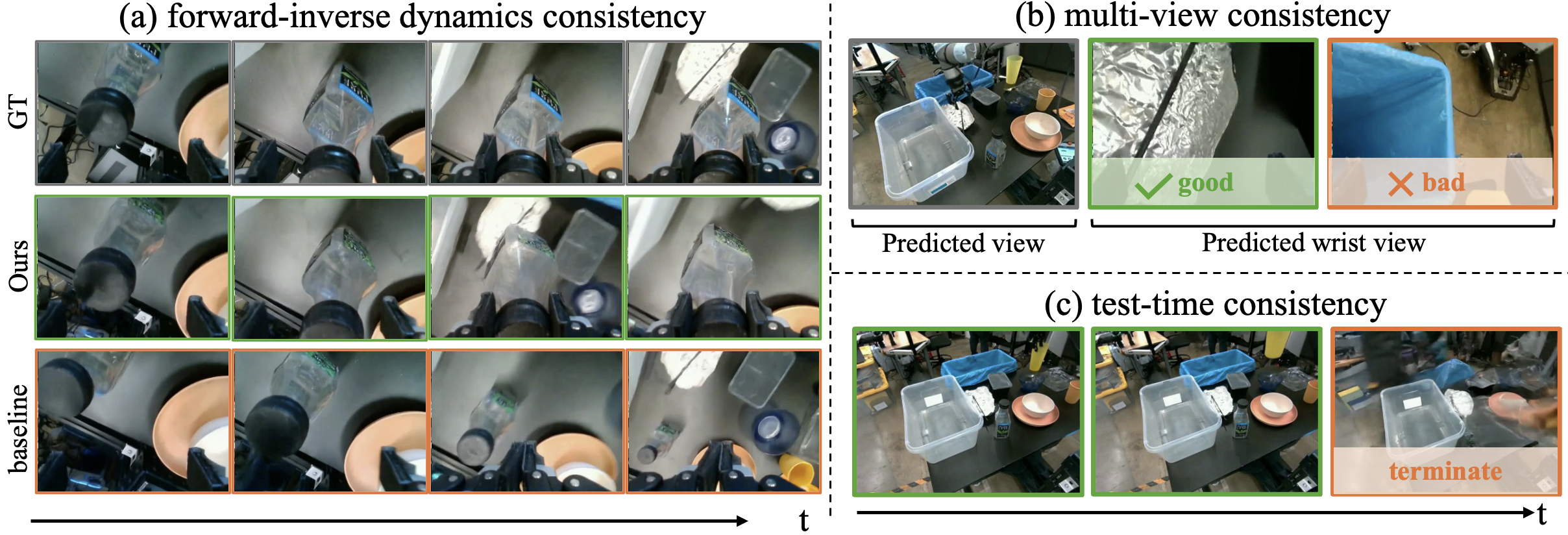}
    \vspace{-16pt}
    \caption{\textbf{The three consistency axes of {\methodname}.} (a) \emph{forward-inverse dynamics consistency.} Under the same action sequence, \emph{Ours} closely tracks \emph{GT}, while a forward-only \emph{baseline} drifts. (b) \emph{multi-view consistency.} The model predicts the wrist view from a \edit{third-person} view, with \emph{good} versus \emph{bad} example predictions. (c) \emph{test-time consistency.} Disagreement between commanded and inverse-mode-recovered actions terminates the rollout once it drifts off-manifold.
    % \TODO{need to upload again}
    }
    \label{fig:pipeline}
    \vspace{-20pt}
\end{figure}

%% file: sources/sec4_related_works.tex
\vspace{-6pt}
\section{\texorpdfstring{\edit{Related Work}}{Related Work}}
\label{sec:related}
\vspace{-6pt}

\paragraph{Real-world policy evaluation.}
Real-world rollouts are the gold standard for evaluating generalist robot policies but are expensive and slow~\citep{atreya2025roboarena}. 
% Scalable alternatives include physics-based simulators~\citep{li2024simpler,yang2026robolab}, real-to-sim approaches~\citep{polaris}, and video-model-based evaluators~\citep{tseng2025scalable,ctrlworld,veorobotics,li2026dworldeval,wang2026interactive}, the last of which we extend.
Scalable alternatives fall into three families. Physics-based simulators such as SIMPLER~\citep{li2024simpler} and RoboLab~\citep{yang2026robolab} require hand-engineered assets and dynamics tuning, leaving residual sim-to-real gaps. Real-to-sim approaches such as PolaRiS~\citep{polaris} reconstruct real scenes via Gaussian splatting but require per-scene reconstruction. Video-model-based evaluators~\citep{tseng2025scalable, ctrlworld, veorobotics, li2026dworldeval, wang2026interactive} forgo asset construction and physics tuning by simulating policy rollouts inside an action-conditioned video model, and this is the family we extend.

\paragraph{Action-conditioned video world models.}
Action-conditioned video models have been studied as world simulators~\citep{irasim, worldgym}. The unified forward-inverse architecture we adopt is from UVA~\citep{li2025uva,zhu2025uwm}, originally proposed as a unified policy and world model for action generation. Concurrent work explores video models as zero-shot policies for visuomotor control and planning~\citep{kim2026cosmos, ye2026worldactionmodelszeroshot} \edit{and exploits the forward-inverse asymmetry to self-verify world-model predictions~\citep{wav26}}, complementing our focus on the \emph{policy evaluation} use case. Our work is among the first to systematically study \edit{the} unified dynamics model for policy evaluation and to identify the inverse dynamics objective as an implicit grounding mechanism that mitigates autoregressive drift. Prior multi-view video models~\citep{ctrlworld, veorobotics} concatenate views and rely on spatial self-attention for cross-view consistency, whereas our multi-view consistency scheme provides explicit supervision by training the model to inpaint missing views from the \edit{remaining} view.

% \paragraph{Long-horizon consistency and memory.}
% Maintaining consistency across long autoregressive rollouts, particularly under occlusion, has been addressed by explicit memory mechanisms: trained memory banks with pose-conditioned retrieval~\citep{worldmem, ctrlworld} and hierarchical history packing~\citep{framepack}. We show that a substantial portion of this failure mode, scene re-entry consistency under occlusion, can be addressed by a much simpler implicit cross-view geometric prior induced by multi-view dropout, with no memory bank, retrieval, or architectural modification beyond the standard forward-inverse training.

\paragraph{Distribution shift and uncertainty in model-based RL.}
Distribution shift between training data and the rollouts a learned dynamics model is queried at is a long-standing challenge in model-based \edit{reinforcement learning (RL)}, where small errors compound autoregressively. Classical responses estimate the model's predictive uncertainty, often through ensemble disagreement~\citep{pets}\edit{, or, in recent video world models, through calibrated generative uncertainty that flags unreliable frames~\citep{wmuncertainty25}}, and then act conservatively when uncertainty is high. MOReL~\citep{morel} converts the uncertainty signal into a pessimistic MDP that terminates rollouts once a threshold is crossed, and MOPO~\citep{mopo} folds it into the reward as a soft penalty during offline policy optimization. Our uncertainty-driven early termination shares the off-manifold detection motivation but differs in two ways. First, the signal is derived from forward-inverse consistency within a single unified world model (Sec.~\ref{sec:method:inference}) rather than from an ensemble, removing the extra training cost. Second, it is applied during policy \emph{evaluation} so that an unreliable rollout is \edit{terminated before further drift contaminates its score}, rather than during offline RL training, where uncertainty must be folded into rewards or used to construct a pessimistic MDP.

%% file: sources/sec2_method.tex
\vspace{-6pt}
\section{SC3-Eval}
\label{sec:method}
\vspace{-6pt}

{\methodname} is a self-consistent video generation recipe that adapts a pre-trained video foundation model into a faithful closed-loop policy evaluator. We first set up the problem and recall the unified dynamics model backbone (Sec.~\ref{sec:method:prelim}). We then describe how training enforces two complementary forms of consistency, forward-inverse dynamics and cross-view (Sec.~\ref{sec:method:training}), and how inference enforces the third, test-time consistency, via an uncertainty-driven early-termination criterion (Sec.~\ref{sec:method:inference}). Implementation details, including the pre-trained weight initialization, training-time augmentations (multi-FPS sampling and pseudo-action augmentation), and optimizer setup, are deferred to App.~\ref{app:training}.

\input{sources/figures/fig_method_combined}
\input{sources/method_01_preliminaries}
\input{sources/method_02_training}

\input{sources/method_03_inference}

%% file: sources/figures/fig_method_combined.tex
\begin{figure}[t]
    \centering
    \begin{minipage}[t]{0.5\textwidth}
        \centering
        \vspace{0pt}
        \includegraphics[width=\linewidth]{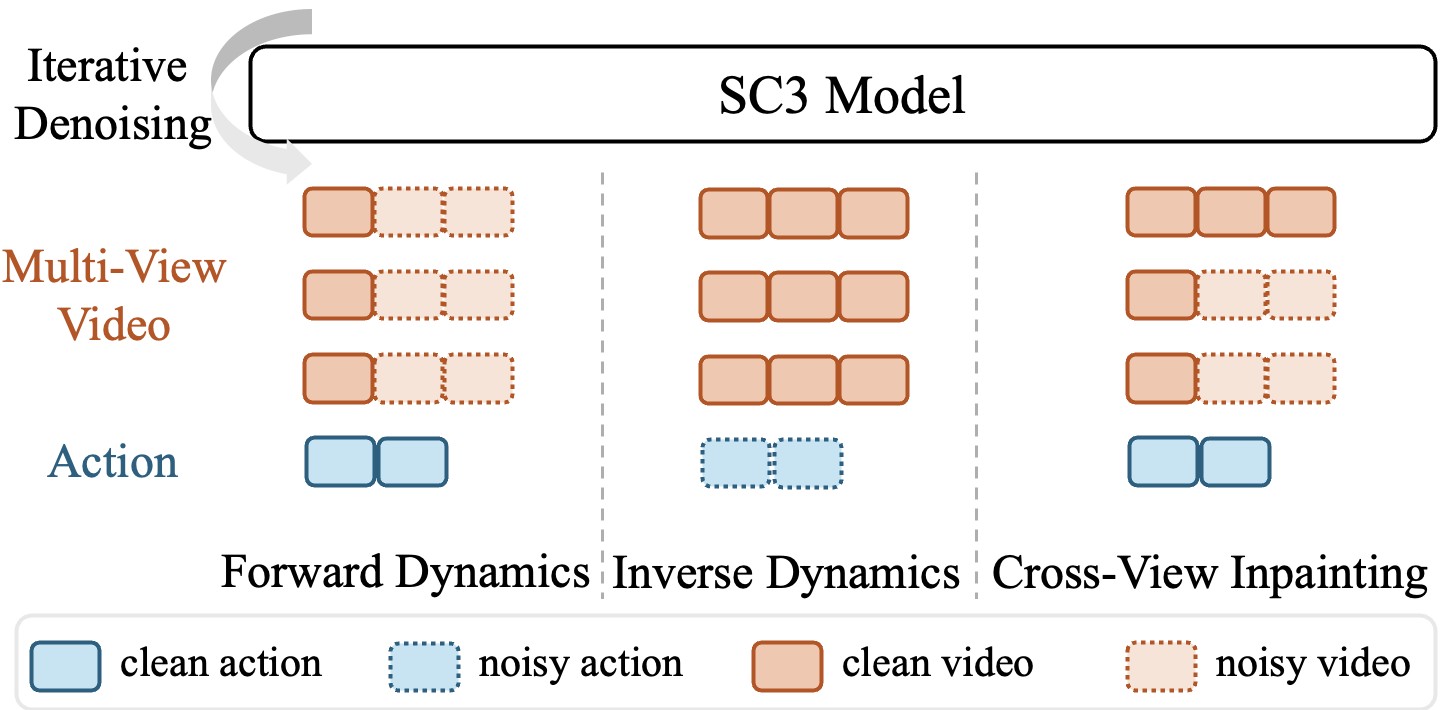}
        \vspace{-12pt}
        % \vspace{-18pt}
        \caption{\textbf{Self-consistency training for \methodname.} Three joint-training modes over a shared backbone, distinguished only by which tokens are clean (conditioning) versus noisy (denoising targets). \emph{Forward Dynamics} denoises future video given action and first frames, \emph{Cross-View Inpainting} denoises held-out views from the \edit{remaining} view and action, and \emph{Inverse Dynamics} denoises the action chunk given the full video. All three modes share parameters and a single flow matching objective.}
        \label{fig:training_mode}
    \end{minipage}\hfill
    \begin{minipage}[t]{0.47\textwidth}
        \centering
        % \vspace{0pt}
        \hrule height 1pt\relax
        \smallskip
        \captionof{algorithm}{Closed-loop policy rollout inside \methodname}
        \vspace{-6pt}
        \label{alg:closed_loop}
        % \smallskip 
        \hrule
        \begin{algorithmic}[1]
        \Require Policy $\pi$, world model with forward mode $\mathcal{W}^{fd}$ and inverse mode $\mathcal{W}^{id}$, initial frame $v_0$, threshold $\tau$, execution horizon $l$, \edit{world-model prediction} horizon $l' > l$
        \State $\mathcal{V} \gets \{v_0\}$
        \For{$t = 0, l, 2l, \dots$}
            \State $a_{t:t+l'} \gets \pi(\mathcal{V})$
            \State $v_{t:t+l'} \gets \mathcal{W}^{fd}(\mathcal{V}, a_{t:t+l'})$
            \State $\hat{a}_{t:t+l} \gets \mathcal{W}^{id}(v_{t:t+l})$
            \State $U_{\mathrm{chunk}}(t) \gets \tfrac{1}{l}\sum_{i=t}^{t+l-1} \| a_i - \hat{a}_i \|_2$
            \If{$U_{\mathrm{chunk}}(t) > \tau$}
                % \State \Return rollout marked as \emph{failed}
                \State \textbf{break}
            \EndIf
            \State $\mathcal{V} \gets \mathcal{V} \cup \{v_{t:t+l}\}$
        \EndFor
        \State \Return scored imagined rollout
        \end{algorithmic}
        \hrule height 1pt\relax
    \end{minipage}
    \vspace{-12pt}
\end{figure}

%% file: sources/method_01_preliminaries.tex
\subsection{Preliminaries}
\label{sec:method:prelim}
\vspace{-6pt}

\paragraph{Problem formulation.}
Let $\Pi = \{\pi_1, \dots, \pi_N\}$ be a set of robot manipulation policies, each with real-world performance $R_i$ measured by the average success rate over a representative task suite under physical rollouts. Our goal is to construct a world simulator $\mathcal{W}$ such that the policy performances $R_{\mathcal{W}, i}$ obtained by rolling out each $\pi_i$ inside $\mathcal{W}$ correlate strongly with the real-world performances $R_i$.
% 
% \paragraph{Faithfulness metrics.}
We quantify faithfulness with two complementary measures. The Pearson correlation coefficient $r(R, R_{\mathcal{W}})$ captures linear agreement between predicted and real-world success rates and measures the \emph{absolute calibration} of the evaluator. The Mean Maximum Rank Violation (MMRV)~\citep{li2024simpler} captures the consistency of pairwise policy rankings, the \emph{relative ordering} property most directly relevant to checkpoint selection in practice.

\paragraph{Unified dynamics model.}
{\methodname} builds on the unified dynamics model architecture, such as UVA~\citep{li2025uva,zhu2025uwm}, a single transformer that operates jointly on video and action tokens. Conventional action-conditioned video models~\citep{irasim,cosmospredict25} learn a single forward map $\mathcal{W}^{fd}(v_{t+1} \mid v_{1:t}, a_{1:t})$, whereas UVA represents video and action in a shared token space and randomly masks subsets of tokens during training. The same network can therefore act as a forward dynamics model that predicts frames from actions, an inverse dynamics model that predicts actions from frames, or a video generator that fills in unobserved frames, depending on which tokens are exposed at inference. We adopt this architecture unchanged and treat it as a black box from here on, with $\mathcal{W}^{fd}$ and $\mathcal{W}^{id}$ denoting its forward and inverse modes.

%% file: sources/method_02_training.tex
\subsection{Self-Consistent Training}
\label{sec:method:training}
\vspace{-6pt}
We adapt the unified dynamics model backbone to multi-view robot manipulation with three jointly trained objectives that enforce two complementary forms of consistency during training, illustrated in Fig.~\ref{fig:training_mode}. The third axis, \emph{test-time consistency}, is realized at inference and described in Sec.~\ref{sec:method:inference}.

\paragraph{Forward-inverse dynamics consistency.}
A forward-only video model lacks a direct mechanism to penalize physically implausible frames, because every frame can be explained as a low-probability rollout under its own distribution. We exploit the model's shared token space and jointly train a forward dynamics mode that reconstructs noised video tokens from the action stream, and an inverse dynamics mode that reconstructs noised action tokens from the video. Because both modes share parameters, the forward mode is implicitly regularized to render frames from which the inverse mode can recover the commanded actions, anchoring generation to a physically plausible action manifold. This forward-inverse parameter sharing acts as an implicit anti-drift regularizer that a separately trained forward-only training mechanism does not provide.

\paragraph{Cross-view consistency.}
Real-world manipulation policies act on multiple synchronized cameras, in our setting two \edit{third-person} cameras and one wrist camera. Without an explicit mechanism, the model tends to drift across views over long autoregressive rollouts, since each view is generated independently from the others. We add a \emph{cross-view inpainting} mode that randomly \edit{selects} one view and asks the model to inpaint the other views, so that each view is supervised to stay consistent with the rest. This implements multi-view consistency without any explicit memory mechanism.

\paragraph{Training procedure.}
We initialize from the pre-trained weights of Cosmos3-Nano~\edit{\citep{cosmos3}} and train the three objectives jointly by random mode assignment, in the same spirit as prior works~\cite{li2025uva,zhu2025uwm}. At each gradient step, every training instance is independently sampled to act as one of the three modes (multi-view forward dynamics, cross-view consistency, or inverse dynamics) with a fixed mixture probability, and the corresponding noised and clean token pattern is applied over the video and action tokens before the transformer is run. Each instance therefore contributes a gradient to a single term of the total objective, and the mode coefficients are realized as the relative sampling frequencies of the three modes rather than as an explicit weighted sum at every step. Because all three modes share the same transformer parameters, the network is supervised to perform forward dynamics, cross-view inpainting, and inverse dynamics within one set of weights. Within each sampled mode, the per-instance loss is a flow matching objective~\citep{lipman2023flow,liu2023flow} on the noised tokens, inheriting the rectified-flow formulation of the Cosmos3 backbone.

%% file: sources/method_03_inference.tex
\vspace{-6pt}
\subsection{Inference with Test-Time Consistency}
\label{sec:method:inference}
\vspace{-6pt}
% \TODO{seems to be not precise enough}
\paragraph{Closed-loop rollout schedule.}
We follow the policy's own receding-horizon schedule. At each step, the policy $\pi$ proposes an action chunk of length $l'$ from the current observation $\mathcal{V}$, the forward mode $\mathcal{W}^{fd}$ renders the corresponding frames, and only the first $l$ frames are appended to $\mathcal{V}$ before re-planning.

\vspace{-6pt}
\paragraph{Prediction-execution horizon decoupling.}
A literal closed-loop simulation would have the world model predict exactly $l$ frames per step~\cite{tseng2025scalable,worldgym}, but this short training horizon is sub-optimal for two distinct reasons. First, \emph{pre-training-prior preservation}. The backbone was pre-trained on substantially longer video clips, so training at a horizon as short as $l$ shifts the training distribution far from that prior and degrades generation quality. Second, \emph{per-chunk supervision richness}. A short $l$-frame chunk exposes the model to too little object behavior per training example to fit the dynamics well, since most short windows contain little inter-frame motion. We address both by training at a longer horizon $l' > l$ and retaining only the first $l$ frames at inference as the next observation context. The extra $l' - l$ frames are discarded at deployment but pay for themselves during training. Alg.~\ref{alg:closed_loop} summarizes the procedure.
% Removed for length: "Our current ablation in Sec.~\ref{sec:exp:ablation} collapses the two effects into a single sweep over $l'$, and disentangling them is left to future work."

\vspace{-6pt}
\paragraph{Uncertainty-driven early termination.}
The same forward-inverse parameter sharing that anchors generated frames during training additionally yields a per-chunk reliability signal at inference. At step $t$, the forward mode $\mathcal{W}^{fd}$ renders frames $v_{t:t+l}$ from the action chunk $a_{t:t+l}$, and the inverse mode $\mathcal{W}^{id}$ recovers an estimate $\hat{a}_{t:t+l}$ from those frames. The per-chunk consistency error
$
U_{\mathrm{chunk}}(t) \;=\; \tfrac{1}{l} \sum_{i=t}^{t+l-1} \left\| a_i - \hat{a}_i \right\|_2
$
measures how well the generated frames remain consistent with the commanded actions. We terminate the rollout once $U_{\mathrm{chunk}}(t) > \tau$, since the forward mode and the action stream have decohered and continuing only compounds the drift. The threshold $\tau$ is chosen on a held-out subset. We read $U_{\mathrm{chunk}}$ as an empirical reliability indicator rather than a calibrated error probability, since the inverse mode sees ground-truth frames during training and generated frames at deployment.

%% file: sources/sec3_experiments.tex
\vspace{-6pt}
\section{Experiments}
\label{sec:experiments}
\vspace{-6pt}

% We evaluate the proposed world model through real-world policy evaluation experiments \TODO{provideo detail in appendix} and study the contribution of its key design choices. Sec.~\ref{sec:exp:setup} describes the experimental setup, evaluation protocol, baselines, and metrics. Sec.~\ref{sec:exp:main} reports the headline result, including a breakdown into in-distribution and out-of-distribution policy checkpoints. Sec.~\ref{sec:exp:ablation} isolates the contribution of inverse dynamics grounding, cross-view grounding, inference-horizon decoupling, and uncertainty-driven early termination through targeted ablations. Sec.~\ref{sec:exp:failure} shows that our evaluator reproduces the specific failure modes of vision-language-action (VLA) policies observed in real-world rollouts, going beyond the standard requirement of matching aggregate success rates. \TODO{If space limited, we can delete this paragraph as well}

\input{sources/experiments_01_setup}
\input{sources/experiments_02_main}
\input{sources/experiments_04_failure}
\input{sources/experiments_03_ablation}

%% file: sources/experiments_01_setup.tex
\subsection{Experimental Setup}
\label{sec:exp:setup}

\paragraph{Dataset and evaluated policies.}
We train the {\methodname} model on a $381$-hour real-world \emph{table bussing} dataset collected by us, and run policy evaluation on this task. To further evaluate the generalization capability of the evaluator, we also run policy evaluation on the \emph{reverse table bussing} task, which is not included in the training dataset.
The workspace and the two task variants are shown in Fig.~\ref{fig:exp_setup}(a). The two tasks share visual appearance and motion primitives but differ in the object-to-destination mapping, so reverse bussing serves as a controlled out-of-distribution task which is not included in the training dataset. 
\edit{In the following}, we refer to the \emph{table bussing} task as the \emph{in-distribution} task and the \emph{reverse table bussing} task as the \emph{out-of-distribution} task.
\edit{Seven VLA policy checkpoints} with the $\pi_{0.5}$~\citep{pi05} architecture are evaluated in our experiments. More details are in App.~\ref{app:data}.
\input{sources/figures/fig_exp_setup}
\paragraph{Evaluation protocol.}
For each policy we collect three matched sets of trajectories under identical initial conditions, \emph{real-world} rollouts from the physical robot, \emph{offline (open-loop)} rollouts that condition the world model on the real-world action sequence (isolating video fidelity from policy interaction), and \emph{online (closed-loop)} rollouts in which the policy acts on the generated frames. Each rollout is observed through three synchronized camera views (Fig.~\ref{fig:exp_setup}(b)) and scored by trained human annotators, blinded to trajectory source, on three independent binary criteria, language following, object lifting, and object placing (Fig.~\ref{fig:exp_setup}(c)). We report Pearson correlation $r$ and Mean Maximum Rank Violation (MMRV)~\citep{li2024simpler}. We compare against three video-model-based evaluators re-run under the same protocol, Ctrl-World~\citep{ctrlworld}, IRASim~\citep{irasim}, and Cosmos-Predict 2.5~\citep{cosmospredict25}. All three support only forward dynamics generation and so cannot enforce the consistency objectives our method does.

%% file: sources/figures/fig_exp_setup.tex
\begin{figure}[t]
    \centering
    \includegraphics[width=\textwidth]{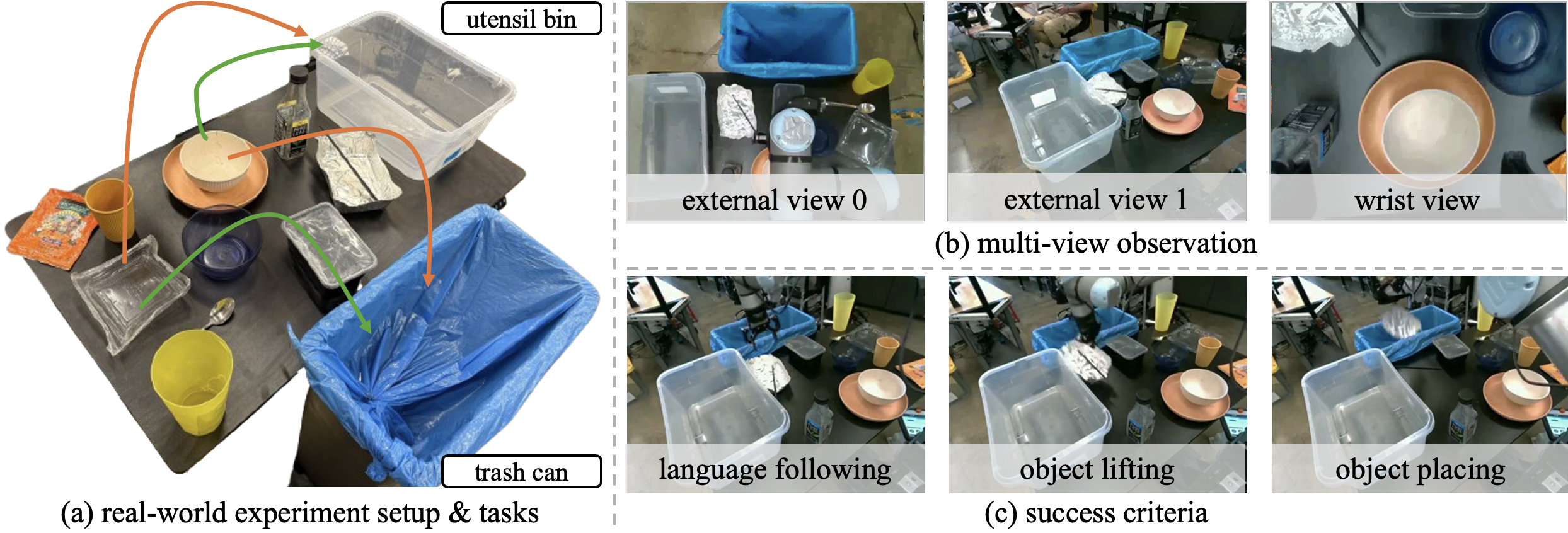}
    \vspace{-12pt}
    \caption{\textbf{Real-world experiment setup.} (a) Workspace, with orange arrows for forward \emph{table bussing} and green arrows for the destination-swapped \emph{reverse} variant. (b) The three synchronized camera views observed by the policy and the world model. (c) Example trajectories scored under the three success criteria.}
    \label{fig:exp_setup}
    \vspace{-6pt}
\end{figure}

%% file: sources/experiments_02_main.tex
\subsection{Evaluating the Policy under \methodname}
\vspace{-6pt}
\label{sec:exp:main}

\input{sources/figures/fig_main_correlation}
\input{sources/tables/tab_main}
\input{sources/figures/fig_qualitative}

Tab.~\ref{tab:main} reports the policy evaluation result, decomposed into in-distribution and out-of-distribution cases, each under offline (open-loop) and online (closed-loop) evaluation. The corresponding correlation plot is shown in Fig.~\ref{fig:main_correlation}.
% 
% \paragraph{Overall faithfulness.}
Our evaluator attains the best MMRV in all four offline/online and in-distribution/out-of-distribution settings, preserving the ranking of policy checkpoints more accurately than every baseline regardless of evaluation mode or task split. It also attains the best Pearson $r$ in three of the four settings. \edit{Evaluators generally degrade} from the in-distribution to the out-of-distribution split, \edit{most sharply in online mode}, reflecting the genuine difficulty of generalizing across a task-semantic shift.
% With only $3$ out-of-distribution checkpoints the absolute out-of-distribution numbers are statistically noisy; we treat them as indicative and view the consistent ranking lead across both splits and both modes as the main takeaway.
\paragraph{Offline versus online.}
One might expect online evaluation to be less faithful than offline evaluation, since the policy must interact with imperfect generated frames. We find that this is not necessarily so. In offline mode the world model is driven by a fixed real-world action sequence, so any generation drift accumulates over the rollout with no corrective feedback. In online mode the closed-loop interaction instead lets the policy react to the generated frames and adjust its commands, partially compensating for drift. Consistent with this, our online faithfulness matches offline on the in-distribution split, with online Pearson $r$ ($0.984$) even exceeding offline ($0.959$). The degradation that does appear is concentrated on the out-of-distribution split, where the world model is itself off-distribution and closed-loop correction cannot fully recover. \edit{These findings argue} for online evaluation as the primary protocol, with the residual gap reflecting out-of-distribution difficulty rather than a closed-loop pathology.

% \paragraph{Score versus success-rate gap.}
% Consistent with prior observations on video-model-based evaluators~\citep{yang2026robolab, veorobotics}, the absolute predicted success rates are systematically lower than ground-truth real-world success rates (see Fig.~\ref{fig:main_correlation}). Critically, however, the \emph{ranking} of checkpoints is preserved with high fidelity, as evidenced by the low MMRV. Since checkpoint selection in practice depends on relative rather than absolute performance, this property is sufficient for the primary use case of the evaluator. We hypothesize that the gap arises because closed-loop rollouts occasionally drift toward visually plausible but task-incorrect states that human annotators score as failures, while the policy on the real robot recovers more often than the world model predicts.

%% file: sources/figures/fig_main_correlation.tex
\begin{figure}[t]
    \centering
    \includegraphics[width=0.95\textwidth]{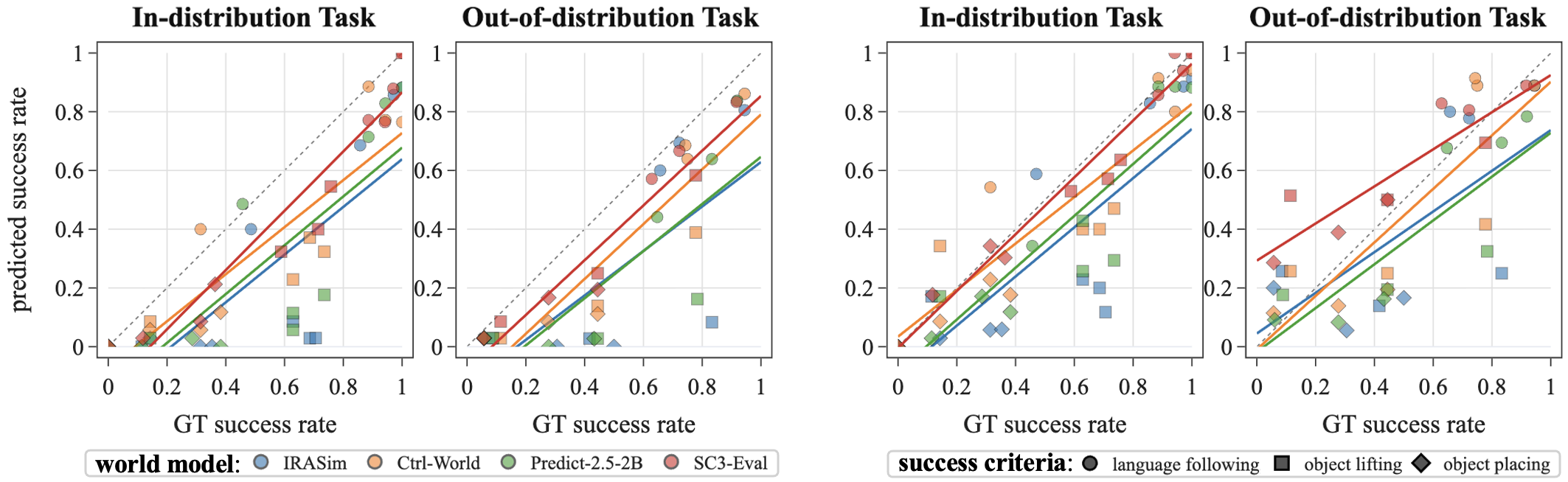}\\[2pt]
    \makebox[0.5\textwidth][c]{\small (a) Offline evaluation (open-loop)}%
    \makebox[0.5\textwidth][c]{\small (b) Online evaluation (closed-loop)}
    \vspace{-12pt}
    \caption{\edit{\textbf{Correlation between predicted and real-world policy performance.}} Each point is a (checkpoint, success-criterion) pair. Within each of (a) and (b), the left and right panels show the in-distribution (table bussing) and out-of-distribution (reverse table bussing) splits.
    % \TODO{legend can be updated}
    }
    \label{fig:main_correlation}
    \vspace{-14pt}
\end{figure}

%% file: sources/tables/tab_main.tex
\begin{table}[t]
\centering
\caption{\edit{\textbf{Policy evaluation results.} Results are split} into in-distribution and out-of-distribution, each evaluated in offline and online mode. Each $r$ and MMRV is computed over the (policy, success criterion) success-rate pairs within a split.
% Pearson $r$: higher is better; MMRV: lower is better.
}
\label{tab:main}
\scriptsize
\renewcommand{\arraystretch}{1.0}
\setlength{\tabcolsep}{4pt}
\begin{tabular*}{\textwidth}{@{\extracolsep{\fill}}lcccccccc@{}}
\toprule
& \multicolumn{4}{c}{In-distribution} & \multicolumn{4}{c}{Out-of-distribution} \\
\cmidrule(lr){2-5} \cmidrule(lr){6-9}
& \multicolumn{2}{c}{Offline} & \multicolumn{2}{c}{Online} & \multicolumn{2}{c}{Offline} & \multicolumn{2}{c}{Online} \\
\cmidrule(lr){2-3} \cmidrule(lr){4-5} \cmidrule(lr){6-7} \cmidrule(lr){8-9}
Method & $r\uparrow$ & MMRV$\downarrow$ & $r\uparrow$ & MMRV$\downarrow$ & $r\uparrow$ & MMRV$\downarrow$ & $r\uparrow$ & MMRV$\downarrow$ \\
\midrule
Ctrl-World~\citep{ctrlworld}              & 0.878 & 0.185 & 0.871 & 0.191 & 0.821 & 0.191 & 0.832 & 0.179 \\
IRASim~\citep{irasim}                     & 0.773 & 0.176 & 0.730 & 0.188 & 0.700 & 0.188 & 0.663 & 0.364 \\
Cosmos-Predict 2.5~\citep{cosmospredict25} & 0.807 & 0.148 & 0.897 & 0.090 & 0.808 & 0.145 & \textbf{0.871} & 0.195 \\
\midrule
\textbf{\methodname{} (Ours)} & \textbf{0.959} & \textbf{0.018} & \textbf{0.984} & \textbf{0.022} & \textbf{0.962} & \textbf{0.022} & 0.870 & \textbf{0.171} \\
\bottomrule
\end{tabular*}
\vspace{-2pt}
\end{table}

%% file: sources/figures/fig_qualitative.tex
\definecolor{legendgreen}{RGB}{86,166,75}
\definecolor{legendblue}{RGB}{47,85,151}
\definecolor{legendorange}{RGB}{224,123,57}
\begin{figure}[t]
    \centering
    \includegraphics[width=\textwidth]{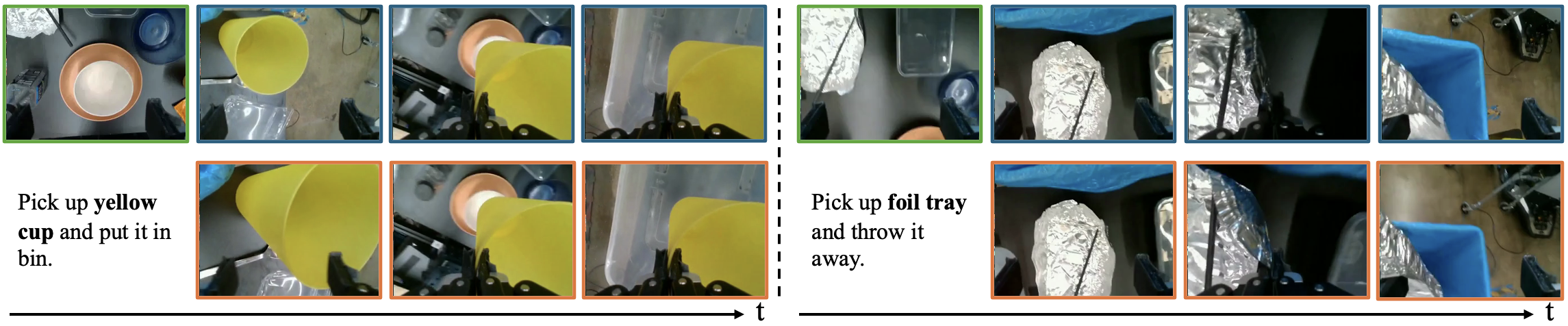}\\[4pt]
    {\small
      \textcolor{legendgreen}{$\blacksquare$}~initial observation \quad
      \textcolor{legendblue}{$\blacksquare$}~real-world rollout \quad
      \textcolor{legendorange}{$\blacksquare$}~online generated rollout
    }
    \vspace{-2pt}
    \caption{\textbf{Qualitative rollouts for \edit{online} generation.} Each example shows the shared initial observation, followed by the real-world rollout and the online rollout generated by our evaluator under the same policy.
    % \TODO{need to find another instance}
    }
    \label{fig:qualitative}
    \vspace{-16pt}
\end{figure}

%% file: sources/experiments_04_failure.tex
\subsection{Reproducing Policy Failure Modes}
\vspace{-6pt}
\label{sec:exp:failure}
\input{sources/figures/fig_failure_repro}
A policy evaluator that reproduces aggregate success rates is useful for checkpoint selection, but a stronger criterion is whether the policy evaluator reproduces the \emph{same kinds of failures} a policy exhibits in the real world. Two evaluators may agree on the success rate yet disagree on which trajectories fail and why. An evaluator that faithfully reproduces behaviors that leads to trajectory failure  is far more reliable  for diagnostic use, such as distinguishing a grasping issue from a perception issue.

Following the three success criteria of Sec.~\ref{sec:exp:setup}, we assign each trajectory to one of four \emph{outcome categories}: \emph{no failure}, or a \emph{language-following}, \emph{object-lifting}, or \emph{object-placing} failure, named by the first criterion it violates. We then ask whether each rollout falls in the same outcome category as its matched real-world trajectory, i.e., whether the evaluator reproduces not just \emph{that} a trajectory fails but the specific failure mode.

Fig.~\ref{fig:failure_reproduction} reports the fraction of real-world trajectories whose matched online (closed-loop) rollouts fall in the same outcome category. Our evaluator attains the highest average reproduction rate and leads in all four categories. \edit{Under closed-loop conditioning, small inconsistencies cascade, so an evaluator that renders plausible but inconsistent frames tends to drifts to a different outcome category than the real-world rollout.} The fact that baselines drop sharply under this protocol while our method holds up motivates the closed-loop setting as the more demanding test of evaluator faithfulness. 
% Offline numbers, where the action sequence is fixed to the real-world action sequence and some baselines score competitively, are reported in App.~\ref{app:abl:failure_table}.

%% file: sources/figures/fig_failure_repro.tex
\begin{wrapfigure}{r}{0.33\textwidth}
    \centering
    \vspace{-\baselineskip}
    \includegraphics[width=\linewidth]{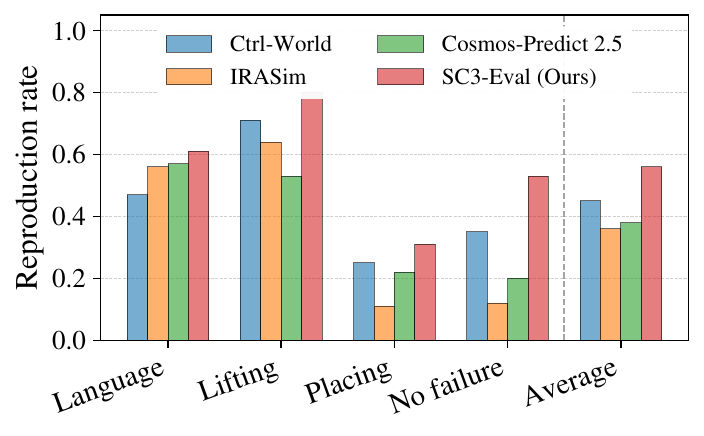}
    \vspace{-12pt}
    \caption{\edit{\textbf{Per-category outcome reproduction rate under online evaluation.} Each bar shows} the fraction of trajectories whose evaluator rollout matches the real-world outcome category.}
    \label{fig:failure_reproduction}
    \vspace{-\baselineskip}
    % \vspace{-18pt}
\end{wrapfigure}

%% file: sources/experiments_03_ablation.tex
\subsection{Ablation Study}
\vspace{-6pt}
\label{sec:exp:ablation}
We isolate the contribution of the key design choices of our evaluator. 
% following the structure of Sec.~\ref{sec:method}: inverse dynamics grounding, cross-view reference forward dynamics, uncertainty-driven early termination, and prediction-execution horizon decoupling. 
Each ablation varies a single factor while holding all other hyperparameters at the values used by the full model. Tab.~\ref{tab:ablations} consolidates the ablations under the same closed-loop policy-level metrics. More results and visualizations are deferred to App.~\ref{app:ablations}.
% we discuss each variant below and defer secondary metrics (offline PSNR, scene re-entry PSNR/LPIPS, in-distribution versus out-of-distribution decomposition) to App.~\ref{app:ablations}.

\vspace{-6pt}
\paragraph{Joint-training modes.}
\input{sources/tables/tab_ablations}
Sec.~\ref{sec:method:training} introduces two grounding training modes, inverse dynamics and \edit{cross-view inpainting}. Removing either degrades both closed-loop metrics (Tab.~\ref{tab:ablations}), confirming that the two mechanisms are complementary rather than redundant. The inverse dynamics ablation \edit{degrades most visibly on the out-of-distribution split in offline visual quality}, where policy commands push rollouts further off the training manifold and the corrective action signal from inverse dynamics matters most. The cross-view ablation primarily hurts wrist-view scene re-entry consistency, indicating that cross-view supervision implicitly preserves scene structure under occlusion, without any explicit memory bank as used by prior work~\citep{ctrlworld, worldmem}. More analysis is deferred to App.~\ref{app:ablations}.
% Full per-split breakdowns, per-step drift curves, and qualitative scene re-entry comparisons are deferred to App.~\ref{app:abl:id} and~\ref{app:abl:cv} and the project website.

\vspace{-6pt}
\paragraph{Uncertainty-driven early termination.}
The per-chunk consistency error $U_{\mathrm{chunk}}$ and the early-termination criterion $U_{\mathrm{chunk}} > \tau$ are defined in Sec.~\ref{sec:method:inference}. Disabling early termination (i.e., letting all rollouts run to completion regardless of consistency error) degrades the closed-loop metrics (Tab.~\ref{tab:ablations})\edit{,} with the largest drop concentrated on the out-of-distribution split where rollout drift is most prevalent.
We also find that the same per-chunk signal additionally serves as context for VLM-based automatic annotation, with details in appendix.

\vspace{-6pt}
\paragraph{Prediction-execution horizon decoupling.}
The full model is trained at a longer prediction horizon than the policy's execution horizon, decoupling the two (Sec.~\ref{sec:method:inference}). Removing this decoupling, i.e., training at the execution horizon directly, degrades the closed-loop metrics (Tab.~\ref{tab:ablations}). Training at such a short horizon shifts the training distribution far from the backbone's much longer video pre-training prior, and each short chunk also exposes the model to too little object behavior to fit the dynamics well.

%% file: sources/tables/tab_ablations.tex
\begin{wraptable}{r}{0.42\textwidth}
\centering
\vspace{-\baselineskip}
\caption{\textbf{Ablation study.} Online closed-loop Pearson and MMRV across \edit{all} policy checkpoints. The first row is our full model; each subsequent row removes (\emph{w/o}) one design choice from Sec.~\ref{sec:method}.}
\label{tab:ablations}
\small
\resizebox{\linewidth}{!}{%
\begin{tabular}{lcc}
\toprule
Variant & Pearson $\uparrow$ & MMRV $\downarrow$ \\
\midrule
\textbf{Full model} & \textbf{0.929} & \textbf{0.119} \\
\midrule
w/o inverse dynamics                  & 0.842 & 0.175 \\
w/o cross-view inpainting     & 0.802 & 0.199 \\
\midrule
w/o early termination    & 0.871 & 0.151 \\
w/o horizon decoupling   & 0.807 & 0.177 \\
\bottomrule
\end{tabular}%
}
% \vspace{-\baselineskip}
% \vspace{-4pt}
\end{wraptable}

%% file: sources/sec5_limitations.tex
\vspace{-6pt}
\section{Limitations}
\label{sec:limitations}
\vspace{-6pt}

\looseness=-1
Closed-loop online evaluation runs at $2.3$ seconds per chunk on a single \edit{GB200} GPU (App.~\ref{app:training:hw}), several orders slower than physics-based simulators~\citep{li2024simpler,yang2026robolab} that run faster than real time, limiting routine checkpoint selection during policy development. Diffusion-sampler acceleration and caching intermediate representations across receding-horizon chunks are natural follow-ups.

\looseness=-1
Our evaluator is also trained and validated on short-horizon manipulation, with a single table-bussing episode lasting roughly $20$ seconds. At substantially longer horizons we expect two failure modes, accumulated drift from a corrective signal that is only local, and visual degradation past the temporal coherence prior of the pre-trained backbone. Extending the evaluator via longer training trajectories or hierarchical sub-goal supervision is a natural next step.

%% file: sources/sec6_conclusions.tex
\vspace{-6pt}
\section{Conclusion}
\vspace{-6pt}
\looseness=-1
We present \methodname, a real-world policy evaluator built by repurposing a pre-trained unified dynamics model. Its joint forward and inverse dynamics, originally designed for action generation, provide much of the structure a faithful evaluator needs. The inverse dynamics objective anchors rollouts to a physically plausible action manifold and counteracts autoregressive drift, while the same mode doubles, at no extra training cost, as a per-chunk uncertainty signal for early termination. Across seven real-world \edit{VLA} policies, {\methodname} achieves a closed-loop Pearson correlation of $0.929$ and MMRV of $0.119$, outperforming three strong prior video-model-based baselines, generalizing to a held-out task semantic, and reproducing per-trajectory failure modes rather than only aggregate success rates. 
% Together with the directions in Sec.~\ref{sec:limitations}, we hope this encourages further research on world-model-based evaluation that treats action grounding as a first-class design principle.

%% file: sources/appendix.tex
% =====================================================================
% Appendix / Supplementary Material
% Sections numbered A, B, ... by the \appendix command in main.tex.
% =====================================================================

\section{Dataset and Evaluation Details}
\label{app:data}

The following appendices give the dataset, training, and ablation details that complement Sec.~\ref{sec:experiments} of the main paper. Full rollout videos and additional qualitative comparisons are hosted on the project website.\footnote{\edit{\projecturl}}

\subsection{Training Dataset}
\label{app:data:pi}

We train \methodname{} on the real-world table bussing dataset introduced in Sec.~\ref{sec:exp:setup}, which spans $381$ hours of robot trajectories collected in a single physical scene with $12$ object categories. Each trajectory provides synchronized streams from three camera views, two \edit{third-person} cameras and one wrist-mounted camera, recorded at $20$~Hz with native resolution $480 \times 640$. We additionally subsample each stream to $10$~Hz to form the multi-FPS training mix described in App.~\ref{app:training:mfps}.

The $12$ object categories cover yellow cup, orange cup, blue bowl, white bowl, orange plate, chopstick, spoon, chip bag, plastic lid, plastic container, foil tray, and bottle. Trajectories cover the table bussing task family (utensils to the utensil bin, trash items to the trash can) and do not include reverse table bussing, which swaps the two destinations and forms the out-of-distribution split used in our evaluation.

Each action is recorded as a delta end-effector pose (delta-EE) and represented as a $7$-dimensional vector with three components for translation, three for axis-angle rotation, and one for gripper width. We normalize each dimension to unit variance using statistics computed over the full training corpus.

\subsection{Evaluation Setup}
\label{app:data:benchmark}

Our real-world evaluation covers seven $\pi_{0.5}$~\citep{pi05} policy checkpoints, four on the in-distribution \emph{table bussing} task and three on the out-of-distribution \emph{reverse table bussing} task. Reverse bussing keeps the same workspace, object set, and language template as table bussing and only swaps the object-to-destination mapping. This isolates the evaluator's generalization to new task semantics from generalization to new pixels or motions. For each checkpoint we collect $36$ to $37$ matched initial conditions, each rolled out both on the physical robot and inside our world-model evaluator from the same starting state for direct comparison. All rollouts, whether physical, offline (open-loop), or online (closed-loop), run for at most $20$ seconds, so that every mode shares the same temporal budget.

\paragraph{Policy checkpoints.}
Tab.~\ref{tab:app_checkpoints} reports the seven checkpoints with their task family and the per-criterion real-world success rates measured in the physical-robot rollouts. The first four checkpoints (IDs 1-4) are in-distribution \emph{table bussing} policies, and the last three (IDs 5-7) are out-of-distribution \emph{reverse table bussing} policies. Each checkpoint predicts $25$ actions at $10$~Hz, and only the first $16$ actions are executed under a receding-horizon strategy before the next replan.

\input{sources/tables/tab_app_checkpoints}

\subsection{Success Criteria and Annotation Protocol}
\label{app:data:annotation}

This subsection details the criteria used to score each rollout and the protocol followed by human annotators.

\paragraph{Success criteria.}
Each rollout is scored on three independent binary criteria described in Sec.~\ref{sec:exp:setup} of the main paper, \emph{language following}, \emph{object lifting}, and \emph{object placing}\edit{, with each criterion scored separately for every object involved in the rollout}. For each checkpoint we compute one success rate per criterion, as the mean of the criterion's \edit{per-object binary scores aggregated over all objects across} the $36$ to $37$ trajectories. Each (checkpoint, criterion) pair then serves as a single data point in the policy-level correlation metrics, giving $7 \times 3 = 21$ data points per evaluation mode.

\paragraph{Policy-level and frame-level metrics.}
At the policy level we report Pearson correlation $r$, which captures linear agreement and absolute calibration, and Mean Maximum Rank Violation (MMRV)~\citep{li2024simpler}, which captures pairwise rank consistency and is the metric most relevant to checkpoint selection. In offline (open-loop) mode the world model is driven by the real-world action sequence, so frame-level fidelity is also meaningful, and we additionally report rollout PSNR averaged over all generated frames. Online (closed-loop) rollouts depart from the real-world frames by construction, so a frame-level reference metric is not meaningful.

\paragraph{Blinding.}
Trajectories from the three sources (real-world, offline open-loop, online closed-loop) are rendered at a uniform $480 \times 640$ resolution and presented to annotators in randomized order. Source labels are stripped from filenames and metadata, and annotators are not told the proportion of trajectories drawn from each source.

\paragraph{Failure-mode taxonomy.}
For each \emph{failed} trajectory, annotators additionally assign exactly one of the three failure modes used in Sec.~\ref{sec:exp:failure} of the main paper, \emph{language-following failure}, \emph{object-lifting failure}, or \emph{object-placing failure}. The taxonomy mirrors the three success criteria, and each failed trajectory is assigned to the first criterion it violates (in the order language $\rightarrow$ lifting $\rightarrow$ placing).

\section{Training Details}
\label{app:training}

\subsection{Network Architecture}
\label{app:training:backbone}

To leverage prior knowledge from vision foundation models, we initialize the network from the pre-trained weights of Cosmos3-Nano, the smallest variant of the Cosmos3 family~\citep{cosmos3}.

Raw delta-EE actions are first normalized per-dimension as described in App.~\ref{app:data:pi}, then projected through an MLP into action embeddings. The resulting embeddings are injected as additive tokens into the time-embedding pathway of each Diffusion Transformer block, and the same action-conditioning pathway is shared between the forward and inverse dynamics objectives.

\subsection{Hardware and Compute}
\label{app:training:hw}

We train \methodname{} on $32$ NVIDIA GB200 GPUs ($8$ nodes with $4$ GPUs per node) for approximately $2.2$ wall-clock days. Inference for closed-loop evaluation runs at $2.3$ seconds per $24$-frame chunk on a single GB200 GPU.

\subsection{Optimization}
\label{app:training:opt}

\paragraph{Optimizer and schedule.}
We use the AdamW optimizer with learning rate $0.0001$, weight decay $0.05$, $\beta_1 = 0.9$, $\beta_2 = 0.99$, and a $2{,}000$-step linear warmup followed by cosine decay. The effective batch size is $512$ trajectories per step ($16$ per-GPU minibatch across $8$ nodes with $4$ GB200 GPUs per node), and we train for $24{,}000$ total steps with the rectified-flow loss inherited from Cosmos3.

\paragraph{Mode mixture.}
At each gradient step, every training instance is independently sampled into one of the three training modes from Sec.~\ref{sec:method:training} with fixed probabilities, $p_{\mathrm{FD}} = 0.8$ for multi-view forward dynamics, $p_{\mathrm{CVI}} = 0.1$ for cross-view inpainting, and $p_{\mathrm{ID}} = 0.1$ for inverse dynamics. The chosen mode determines the clean/noisy partition over video and action tokens before the forward pass, so each instance contributes a gradient to a single mode, and the relative mode weights are realized as these sampling frequencies rather than as an explicit weighted sum at every step.

\subsection{Training Augmentations}
\label{app:training:aug}

\paragraph{Pseudo-action augmentation.}
\label{app:training:pseudo}
With probability $p_{\mathrm{pseudo}} = 0.5$, the policy command for a minibatch sample is replaced by a pseudo-action computed from consecutive real-world end-effector poses. The pseudo-action and policy-command distributions occupy slightly different action manifolds, and exposing the model to both during training provides marginal robustness gains.

\paragraph{Multi-FPS training.}
\label{app:training:mfps}
We jointly train at $10$ and $20$ FPS by sampling the temporal stride uniformly per minibatch. The two FPS settings correspond to effective action horizons of $2.4$ and $1.2$ seconds for the $24$-step chunks used during training. They expose the model to complementary temporal statistics within a fixed chunk budget. The $20$~FPS stream provides fine-grained inter-frame information that is useful for short-range physics and contact, while the $10$~FPS stream covers twice the temporal span and exposes the model to more global movement per chunk. Since the transformer does not directly observe the frame rate through its positional encoding, this is best read as a data-augmentation effect that broadens the visual statistics each chunk represents, rather than as an explicit timescale-conditioning signal.

\subsection{Inference Procedure}
\label{app:training:inference}

\paragraph{Prediction-execution horizon decoupling.}
Following the notation of Sec.~\ref{sec:method:inference} of the main paper, the world model is trained at a prediction horizon of $l' = 24$ frames, of which only the first $l = 16$ frames are retained at inference, matching the policy's $16$-step receding-horizon execution. Training directly at $l = 16$ frames degrades generation quality, since this horizon is far shorter than the $400$-frame video clips on which Cosmos3 was pre-trained. At each closed-loop step the policy proposes a $25$-action chunk, of which we feed the first $l' = 24$ actions to the world model. The world model generates $l'$ future frames in each of the three camera views, and we keep the first $l$ frames per view as the observation context for the next chunk.

\paragraph{Early-termination threshold.}
For the uncertainty-driven early-termination criterion used in Sec.~\ref{sec:exp:ablation}, we tune the per-chunk threshold to $\tau = 0.02$ on a held-out subset of $30$ trajectories. This threshold terminates approximately $4.88\%$ of rollouts. \edit{Performance is stable for $\tau \in [0.02, 0.05]$ and degrades under more aggressive termination at $\tau = 0.01$ (App.~\ref{app:abl:uncertainty}).}

\section{Additional Ablation Study}
\label{app:ablations}

The main paper evaluates each design choice via closed-loop policy-level correlation metrics in Tab.~\ref{tab:ablations}. This appendix complements that analysis by examining how each design factor affects the visual quality of offline (open-loop) generation. Tab.~\ref{tab:app_psnr_offline} reports rollout PSNR against the real-world frames per method and per ablation variant. Online (closed-loop) rollouts depart from the real-world frames by construction, so PSNR is meaningful only in offline mode. Subsequent subsections analyze each design factor in turn\edit{, and App.~\ref{app:abl:uncertainty} additionally provides a finer InD/OOD breakdown of the closed-loop metrics for the early-termination ablation}.

\input{sources/tables/tab_app_psnr_offline}

\subsection{Training Modes}
\label{app:abl:training_modes}

\methodname{} introduces two grounding training modes alongside multi-view forward dynamics (Sec.~\ref{sec:method:training}), inverse dynamics and cross-view inpainting. We ablate each in turn below.

\paragraph{Inverse dynamics.}
\label{app:abl:id}
Removing the inverse dynamics grounding objective degrades offline visual quality on both splits (Tab.~\ref{tab:app_psnr_offline}, w/o inverse dynamics row), and the OOD drop ($1.05$ PSNR) is more than twice the InD drop ($0.47$ PSNR).
% Removed (review item 2.2): "The asymmetry is also visible across the ablation row, where w/o inverse dynamics produces the lowest OOD PSNR of any variant and is barely above the strongest video-world-model baseline."
The corrective signal from inverse dynamics becomes most valuable when policy commands drive the rollout off the training-data manifold, where the rollout otherwise drifts without correction. Fig.~\ref{fig:qual_id_ablation} illustrates this qualitatively on a single offline trajectory, where the variant without inverse dynamics tracks the ground truth for the first second and then diverges onto a different scene, while the variant with inverse dynamics follows the bottle pick-and-place motion through the end of the rollout.

\input{sources/figures/fig_qual_id_ablation}

\paragraph{Cross-view inpainting.}
\label{app:abl:cv}
The cross-view inpainting ablation has a particularly striking qualitative effect on wrist-view scene re-entry. When the wrist camera temporarily turns away from the workspace and later returns, the variant trained without cross-view inpainting hallucinates the workspace and never recovers once it becomes visible again, while the variant trained with cross-view inpainting restores the correct workspace state. We identify re-entry segments automatically as rollout windows in which the wrist camera workspace visibility transitions from below threshold to above threshold. Fig.~\ref{fig:qual_cv_ablation} compares both variants over the same re-entry segment at four timesteps, with the zoomed wrist view on top and the two \edit{third-person} views below for each panel.

\input{sources/figures/fig_qual_cv_ablation}

\subsection{Prediction-Execution Horizon Decoupling}
\label{app:abl:horizon}

The prediction-execution horizon decoupling result in Tab.~\ref{tab:ablations} of the main paper carries over to the offline (open-loop) mode, with the w/o horizon decoupling row of Tab.~\ref{tab:app_psnr_offline} showing a consistent PSNR drop. Setting $l' = l$ (no decoupling) shortens the per-chunk video horizon below the world model's pre-training prior, and degrades both closed-loop policy-level correlation and offline visual quality.

\subsection{Uncertainty-Driven Early Termination}
\label{app:abl:uncertainty}

\input{sources/figures/fig_uncertainty}

Tab.~\ref{tab:app_uncertainty} reports the uncertainty-driven early-termination ablation split by InD and OOD, together with the termination rate (fraction of rollouts terminated by the criterion) at several threshold settings.

\input{sources/tables/tab_app_uncertainty}

\paragraph{Asymmetry between InD and OOD.}
\edit{Early termination yields consistent gains} on the OOD split, with \edit{every threshold setting improving over the no-termination baseline and} OOD~$r$ rising from $0.853$ at the \edit{no-termination} baseline to $0.902$ at the most aggressive setting $\tau_3 = 0.01$, and OOD~MMRV trending downward over the same sweep ($0.197 \rightarrow 0.159$). \edit{On the InD split the gains saturate earlier: both metrics are best at $\tau_2 = 0.02$ and degrade under more aggressive termination.} This asymmetry is consistent with the interpretation that the inverse-dynamics consistency error $U_{\mathrm{chunk}}$ most often flags rollouts that have drifted away from the training manifold, and such drift is more frequent on OOD prompts that push the policy off the manifold to begin with.

\paragraph{Sweet spot at $\tau_2 = 0.02$.}
The InD result is non-monotonic in $\tau$. Tightening the threshold from $\tau_2 = 0.02$ to $\tau_3 = 0.01$ raises the termination rate from \edit{$4.88\%$} to $10.22\%$ but degrades InD~$r$ from \edit{$0.984$} to $0.941$ and InD~MMRV from \edit{$0.022$} to $0.101$. The drop is sharp enough to indicate that the additional rollouts \edit{terminated} at $\tau_3$ are predominantly rollouts the world model would have continued correctly, so \edit{truncating them turns would-be successes into scored failures} faster than it improves per-rollout fidelity. We therefore adopt $\tau_2 = 0.02$ as our operating point, which captures \edit{part} of the OOD gain (\edit{$0.870$} versus the $0.902$ ceiling) while sitting at the InD optimum.

\paragraph{$U_{\mathrm{chunk}}$ for VLM annotation.}
\label{app:abl:vlm}
Beyond early termination, the per-chunk consistency error $U_{\mathrm{chunk}}$ is also useful as side information for automatic rollout scoring. While our main results rely on human annotators, we additionally evaluate whether a vision-language model (VLM, specifically Gemini 3.1-pro) can substitute for human raters when given $U_{\mathrm{chunk}}$ as additional context. Without $U_{\mathrm{chunk}}$, the VLM-derived policy-level success rate is $0.712$. Supplying $U_{\mathrm{chunk}}$ values alongside the rendered frames raises it to $0.761$, closer to the human-annotated number used in the main paper.

\subsection{Per-Category Outcome Reproduction}
\label{app:abl:failure}
\label{app:abl:failure_table}

Fig.~\ref{fig:failure_reproduction} of the main paper visualizes the per-category outcome reproduction rates as a bar plot. Tab.~\ref{tab:failure_reproduction} reports the same numbers in tabular form for reference, including the offline (open-loop) column alongside the online (closed-loop) column shown in the main paper.

\input{sources/tables/tab_failure_reproduction}

The offline column is relatively easy for every method, since each rollout is conditioned on the real-world action sequence and therefore largely retraces the same trajectory. The online column is the harder and more informative test, where the policy acts on generated frames and small inconsistencies compound. Under this protocol, baselines drop sharply across categories, while \methodname{} retains the highest reproduction rate on every category.

% \bibliography{example}  % .bib

%% file: sources/tables/tab_app_checkpoints.tex
\begin{table}[h]
\centering
\caption{\edit{\textbf{Policy checkpoints used for evaluation.} Each row reports the} real-world success rate per success criterion.}
\label{tab:app_checkpoints}
\small
\setlength{\tabcolsep}{4pt}
\begin{tabular*}{\linewidth}{@{\extracolsep{\fill}}llccc@{}}
\toprule
 & & \multicolumn{3}{c}{Real-world success rate} \\
\cmidrule(lr){3-5}
ID & Task family       & Language following & Object lifting & Object placing \\
\midrule
$1$ & Table bussing (InD)         & $0.946$ & $0.610$ & $0.126$ \\
$2$ & Table bussing (InD)         & $0.876$ & $0.688$ & $0.306$ \\
$3$ & Table bussing (InD)         & $0.959$ & $0.720$ & $0.350$ \\
$4$ & Table bussing (InD)         & $0.446$ & $0.154$ & $0.008$ \\
\midrule
$5$ & Reverse table bussing (OOD) & $0.745$ & $0.444$ & $0.286$ \\
$6$ & Reverse table bussing (OOD) & $0.675$ & $0.105$ & $0.057$ \\
$7$ & Reverse table bussing (OOD) & $0.926$ & $0.785$ & $0.449$ \\
\bottomrule
\end{tabular*}
\end{table}

%% file: sources/tables/tab_app_psnr_offline.tex
\begin{table}[t]
\centering
\caption{\edit{\textbf{Offline rollout PSNR per method and per ablation variant.} Results are split} into in-distribution (table bussing) and out-of-distribution (reverse table bussing). Higher is better.}
\label{tab:app_psnr_offline}
\footnotesize
\renewcommand{\arraystretch}{1.0}
\setlength{\tabcolsep}{4pt}
\begin{tabular*}{\linewidth}{@{\extracolsep{\fill}}lcc@{}}
\toprule
Method & In-distribution PSNR$\uparrow$ & Out-of-distribution PSNR$\uparrow$ \\
\midrule
Ctrl-World~\citep{ctrlworld}               & 14.80 & 14.61 \\
IRASim~\citep{irasim}                      & 13.99 & 13.12 \\
Cosmos-Predict 2.5~\citep{cosmospredict25} & 14.83 & 14.81 \\
\midrule
w/o inverse dynamics       & 14.97 & 14.07 \\
w/o cross-view inpainting  & 15.11 & 15.01 \\
w/o horizon decoupling     & 14.87 & 14.76 \\
\midrule
\textbf{\methodname{} (Ours)} & \textbf{15.44} & \textbf{15.12} \\
\bottomrule
\end{tabular*}
\vspace{-12pt}
\end{table}

%% file: sources/figures/fig_qual_id_ablation.tex
\definecolor{idblue}{RGB}{47,85,151}
\definecolor{idgreen}{RGB}{86,166,75}
\definecolor{idorange}{RGB}{224,123,57}
\begin{figure}[t]
    \centering
    \includegraphics[width=\textwidth]{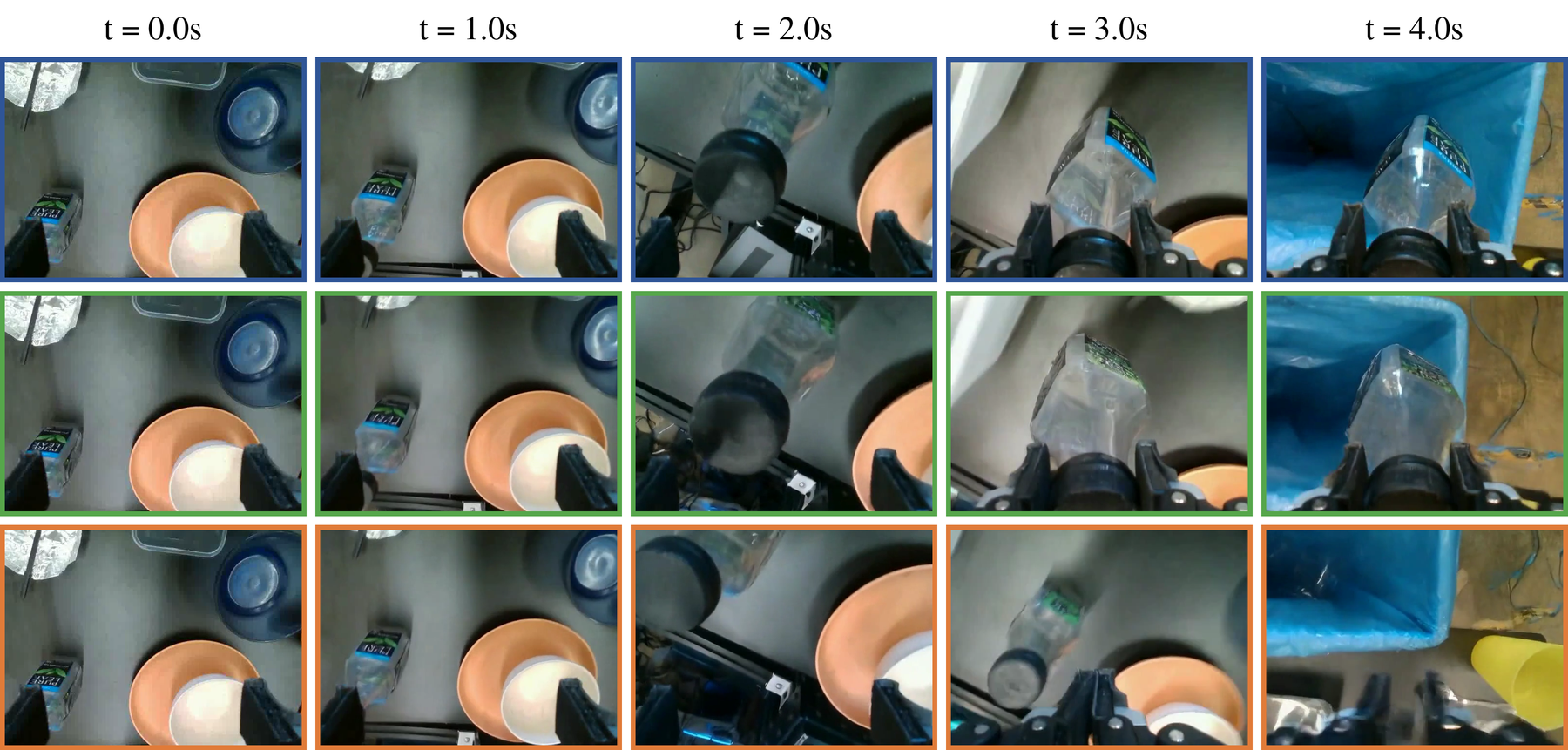}\\[4pt]
    {\small
      \textcolor{idblue}{$\blacksquare$}~Ground truth \quad
      \textcolor{idgreen}{$\blacksquare$}~w/ inverse dynamics \quad
      \textcolor{idorange}{$\blacksquare$}~w/o inverse dynamics
    }
    \vspace{-6pt}
    \caption{\textbf{Qualitative effect of inverse dynamics grounding on offline rollouts.} Each row shows the same offline (open-loop) rollout under a different training configuration, sampled at five timesteps from a single bussing trajectory. The variant trained without inverse dynamics matches the ground truth for the first second, then drifts onto a different scene and never recovers, while the variant with inverse dynamics tracks the ground-truth bottle pick-and-place motion through the end of the rollout.}
    \label{fig:qual_id_ablation}
    \vspace{-12pt}
\end{figure}

%% file: sources/figures/fig_qual_cv_ablation.tex
\definecolor{cvblue}{RGB}{47,85,151}
\definecolor{cvgreen}{RGB}{86,166,75}
\begin{figure}[t]
    \centering
    \includegraphics[width=\textwidth]{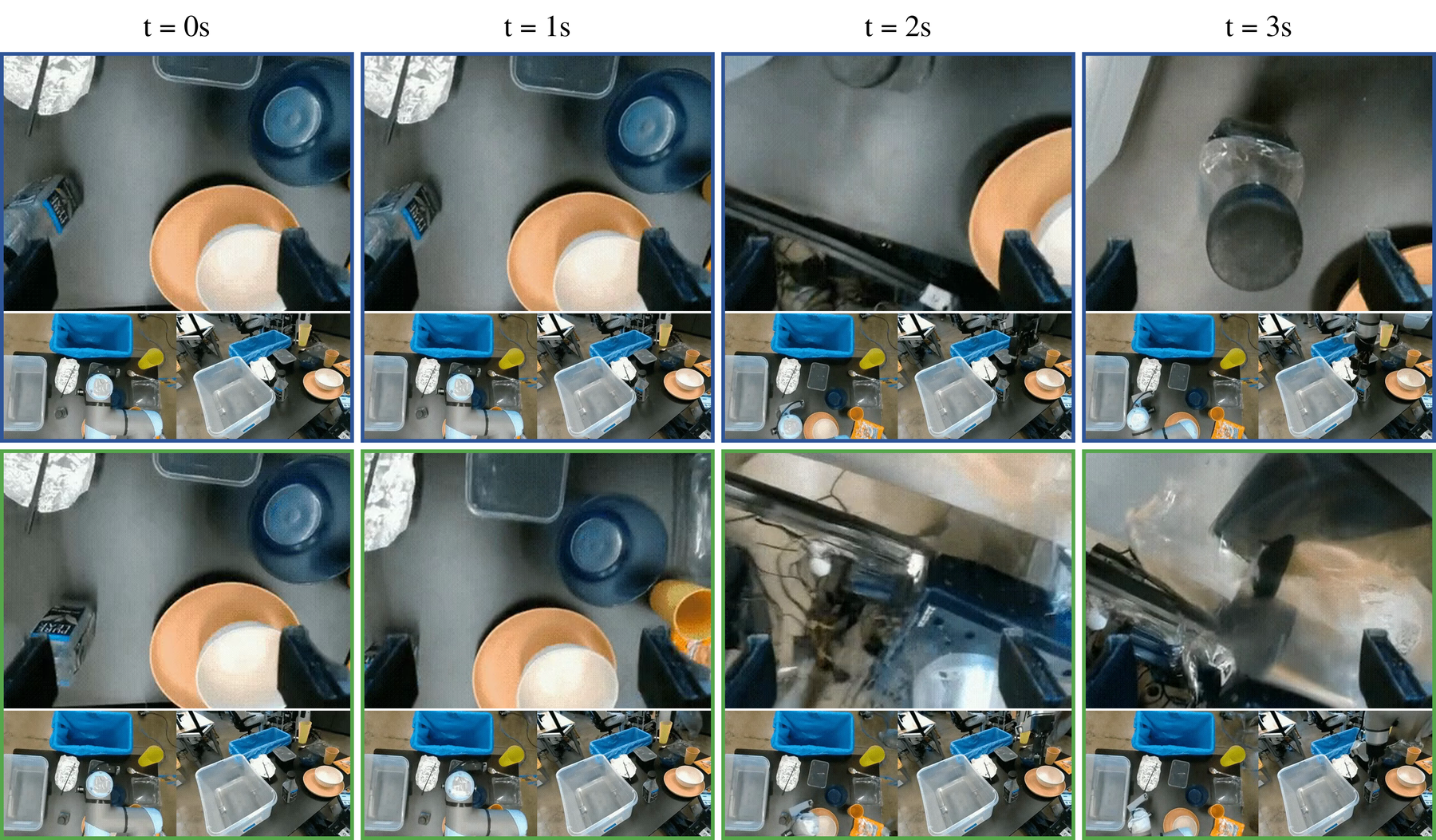}\\[4pt]
    {\small
      \textcolor{cvblue}{$\blacksquare$}~w/ cross-view inpainting \quad
      \textcolor{cvgreen}{$\blacksquare$}~w/o cross-view inpainting
    }
    \vspace{-6pt}
    \caption{\textbf{Qualitative effect of cross-view inpainting on wrist-view scene re-entry.} Each row shows the same offline (open-loop) rollout under a different training configuration, sampled at four timesteps spanning a re-entry segment in which the wrist camera turns away from the workspace and then returns. Each panel shows a zoomed wrist view on top and the two \edit{third-person} views below. The variant trained without cross-view inpainting hallucinates the wrist content and never resolves once the workspace becomes visible again, while the variant trained with cross-view inpainting recovers the correct workspace state under occlusion.}
    \label{fig:qual_cv_ablation}
    \vspace{-12pt}
\end{figure}

%% file: sources/figures/fig_uncertainty.tex
\begin{figure}[t]
    \centering
    \includegraphics[width=\textwidth]{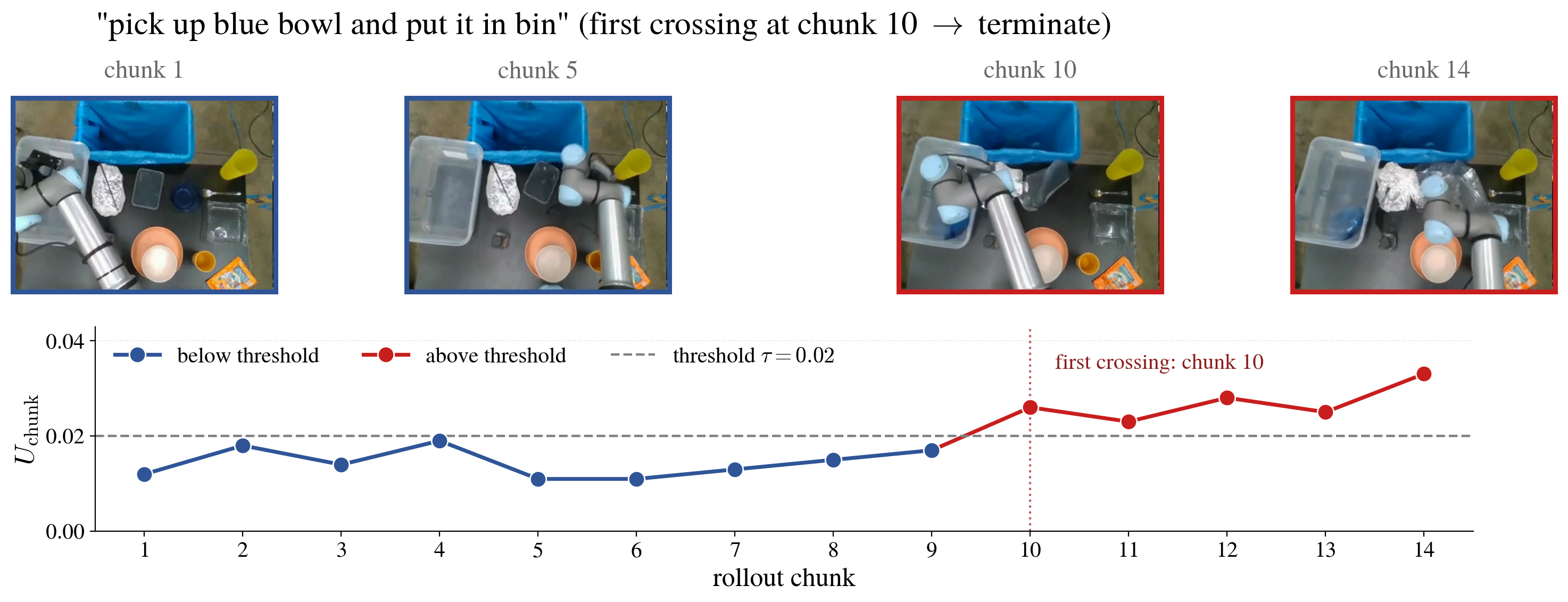}
    \vspace{-8pt}
    \caption{\edit{\textbf{Uncertainty-driven early termination on an example rollout.}} Top, four frames sampled from the imagined rollout. Bottom, the per-chunk consistency error $U_{\mathrm{chunk}}$ over the $14$ rollout chunks. The rollout is terminated at the first chunk whose $U_{\mathrm{chunk}}$ exceeds the threshold $\tau = 0.02$ (chunk $10$ here), before the visible drift compounds.}
    \label{fig:uncertainty}
    \vspace{-8pt}
\end{figure}

%% file: sources/tables/tab_app_uncertainty.tex
\begin{table}[t]
\centering
\caption{\edit{\textbf{Uncertainty-driven early termination ablation.} Results are decomposed} by InD/OOD and shown across threshold settings.}
\label{tab:app_uncertainty}
\footnotesize
\renewcommand{\arraystretch}{1.0}
\setlength{\tabcolsep}{4pt}
\begin{tabular*}{\linewidth}{@{\extracolsep{\fill}}lccccc@{}}
\toprule
& & \multicolumn{2}{c}{In-distribution} & \multicolumn{2}{c}{Out-of-distribution} \\
\cmidrule(lr){3-4} \cmidrule(lr){5-6}
Threshold $\tau$ & Termination rate & $r\uparrow$ & MMRV$\downarrow$ & $r\uparrow$ & MMRV$\downarrow$ \\
\midrule
$\infty$ (\edit{no termination}) & $0\%$    & 0.956          & 0.093          & 0.853          & 0.197 \\
$\tau_1 = 0.05$      & $2.21\%$  & 0.974          & 0.027          & 0.881          & 0.170 \\
$\tau_2 = 0.02$      & \edit{$4.88\%$} & \edit{\textbf{0.984}} & \edit{\textbf{0.022}} & \edit{0.870} & \edit{0.171} \\
$\tau_3 = 0.01$      & $10.22\%$ & 0.941          & 0.101          & \edit{\textbf{0.902}} & \edit{\textbf{0.159}} \\
\bottomrule
\end{tabular*}
\vspace{-12pt}
\end{table}

%% file: sources/tables/tab_failure_reproduction.tex
\begin{table}[t]
\centering
\caption{\edit{\textbf{Outcome reproduction rate per category, reported as \emph{offline / online}.}} Each cell shows the fraction of real-world trajectories whose matched offline (open-loop) and online (closed-loop) rollouts have the same outcome category as the real-world rollout. Higher is better in both modes.}
\label{tab:failure_reproduction}
\small
\setlength{\tabcolsep}{4pt}
\begin{tabular*}{\linewidth}{@{\extracolsep{\fill}}l r@{\,/\,}l r@{\,/\,}l r@{\,/\,}l r@{\,/\,}l r@{\,/\,}l@{}}
\toprule
& \multicolumn{10}{c}{Per-category reproduction rate (offline / online)} \\
\cmidrule(lr){2-11}
Method & \multicolumn{2}{c}{Language} & \multicolumn{2}{c}{Lifting} & \multicolumn{2}{c}{Placing} & \multicolumn{2}{c}{No failure} & \multicolumn{2}{c}{Average $\uparrow$} \\
\midrule
Ctrl-World~\citep{ctrlworld}                & 0.25 & 0.47 & 0.90 & 0.71 & \textbf{0.81} & 0.25 & 0.25 & 0.35 & 0.55 & 0.45 \\
IRASim~\citep{irasim}                       & 0.03 & 0.56 & \textbf{0.92} & 0.64 & 0.80 & 0.11 & 0.05 & 0.12 & 0.45 & 0.36 \\
Cosmos-Predict 2.5~\citep{cosmospredict25}  & \textbf{0.91} & 0.57 & 0.77 & 0.53 & 0.15 & 0.22 & 0.02 & 0.20 & 0.46 & 0.38 \\
\midrule
\textbf{Ours} & \textbf{0.91} & \textbf{0.61} & 0.72 & \textbf{0.80} & 0.47 & \textbf{0.31} & \textbf{0.38} & \textbf{0.53} & \textbf{0.62} & \textbf{0.56} \\
\bottomrule
\end{tabular*}
% \vspace{-12pt}
\end{table}

%% file: example.bib
@article{cosmospredict25,
  title   = {World Simulation with Video Foundation Models for Physical AI},
  author  = {NVIDIA and Ali, Arslan and Bai, Junjie and Bala, Maciej and Balaji, Yogesh and Blakeman, Aaron and Cai, Tiffany and Cao, Jiaxin and Cao, Tianshi and Cha, Elizabeth and Chao, Yu-Wei and Chattopadhyay, Prithvijit and Chen, Mike and Chen, Yongxin and Chen, Yu and Cheng, Shuai and Cui, Yin and Diamond, Jenna and Ding, Yifan and Fan, Jiaojiao and Fan, Linxi and Feng, Liang and Ferroni, Francesco and Fidler, Sanja and Fu, Xiao and Gao, Ruiyuan and Ge, Yunhao and Gu, Jinwei and Gupta, Aryaman and Gururani, Siddharth and El Hanafi, Imad and Hassani, Ali and Hao, Zekun and Huffman, Jacob and Jang, Joel and Jannaty, Pooya and Kautz, Jan and Lam, Grace and Li, Xuan and Li, Zhaoshuo and Liao, Maosheng and Lin, Chen-Hsuan and Lin, Tsung-Yi and Lin, Yen-Chen and Ling, Huan and Liu, Ming-Yu and Liu, Xian and Lu, Yifan and Luo, Alice and Ma, Qianli and Mao, Hanzi and Mo, Kaichun and Nah, Seungjun and Narang, Yashraj and Panaskar, Abhijeet and Pavao, Lindsey and Pham, Trung and Ramezanali, Morteza and Reda, Fitsum and Reed, Scott and Ren, Xuanchi and Shao, Haonan and Shen, Yue and Shi, Stella and Song, Shuran and Stefaniak, Bartosz and Sun, Shangkun and Tang, Shitao and Tasmeen, Sameena and Tchapmi, Lyne and Tseng, Wei-Cheng and Varghese, Jibin and Wang, Andrew Z. and Wang, Hao and Wang, Haoxiang and Wang, Heng and Wang, Ting-Chun and Wei, Fangyin and Xu, Jiashu and Yang, Dinghao and Yang, Xiaodong and Ye, Haotian and Ye, Seonghyeon and Zeng, Xiaohui and Zhang, Jing and Zhang, Qinsheng and Zheng, Kaiwen and Zhu, Andrew and Zhu, Yuke},
  journal = {arXiv preprint arXiv:2511.00062},
  year    = {2025}
}

@inproceedings{atreya2025roboarena,
  title     = {RoboArena: Distributed Real-World Evaluation of Generalist Robot Policies},
  author    = {Atreya, Pranav and Pertsch, Karl and Lee, Tony and Kim, Moo Jin and Jain, Arhan and Kuramshin, Artur and Eppner, Clemens and Neary, Cyrus and Hu, Edward and Ramos, Fabio and others},
  booktitle = {Proceedings of the Conference on Robot Learning (CoRL 2025)},
  year      = {2025}
}

@article{li2024simpler,
  title   = {Evaluating Real-World Robot Manipulation Policies in Simulation},
  author  = {Xuanlin Li and Kyle Hsu and Jiayuan Gu and Karl Pertsch and Oier Mees and Homer Rich Walke and Chuyuan Fu and Ishikaa Lunawat and Isabel Sieh and Sean Kirmani and Sergey Levine and Jiajun Wu and Chelsea Finn and Hao Su and Quan Vuong and Ted Xiao},
  journal = {arXiv preprint arXiv:2405.05941},
  year    = {2024}
}

@misc{yang2026robolab,
  title         = {RoboLab: A High-Fidelity Simulation Benchmark for Analysis of Task Generalist Policies},
  author        = {Xuning Yang and Rishit Dagli and Alex Zook and Hugo Hadfield and Ankit Goyal and Stan Birchfield and Fabio Ramos and Jonathan Tremblay},
  year          = {2026},
  url           = {https://arxiv.org/abs/2604.09860}
}

@misc{polaris,
  title   = {PolaRiS: Scalable Real-to-Sim Evaluations for Generalist Robot Policies},
  author  = {Jain, Arhan and Zhang, Mingtong and Arora, Kanav and Chen, William and Torne, Marcel and Irshad, Muhammad Zubair and Zakharov, Sergey and Wang, Yue and Levine, Sergey and Finn, Chelsea and Ma, Wei-Chiu and Shah, Dhruv and Gupta, Abhishek and Pertsch, Karl},
  year    = {2025},
  url     = {https://arxiv.org/abs/2512.16881}
}

@article{tseng2025scalable,
  title   = {Scalable Policy Evaluation with Video World Models},
  author  = {Tseng, Wei-Cheng and Gu, Jinwei and Zhang, Qinsheng and Mao, Hanzi and Liu, Ming-Yu and Shkurti, Florian and Lin, Yen-Chen},
  journal = {arXiv preprint arXiv:2511.11520},
  year    = {2025},
  url     = {https://arxiv.org/abs/2511.11520}
}

@article{veorobotics,
  title   = {Evaluating Gemini Robotics Policies in a Veo World Simulator},
  author  = {{Gemini Robotics Team}},
  journal = {arXiv preprint arXiv:2512.10675},
  year    = {2025},
  url     = {https://arxiv.org/abs/2512.10675}
}

@article{ctrlworld,
  title   = {Ctrl-World: A Controllable Generative World Model for Robot Manipulation},
  author  = {Guo, Yanjiang and Shi, Lucy Xiaoyang and Chen, Jianyu and Finn, Chelsea},
  journal = {arXiv preprint arXiv:2510.10125},
  year    = {2025},
  url     = {https://arxiv.org/abs/2510.10125}
}

@article{worldgym,
  title   = {WorldGym: World Model as An Environment for Policy Evaluation},
  author  = {Quevedo, Julian and Sharma, Ansh Kumar and Sun, Yixiang and Suryavanshi, Varad and Liang, Percy and Yang, Sherry},
  journal = {arXiv preprint arXiv:2506.00613},
  year    = {2025},
  doi     = {10.48550/arXiv.2506.00613},
  url     = {https://arxiv.org/abs/2506.00613}
}

@inproceedings{irasim,
  title     = {IRASim: A Fine-Grained World Model for Robot Manipulation},
  author    = {Zhu, Fangqi and Wu, Hongtao and Guo, Song and Liu, Yuxiao and Cheang, Chilam and Kong, Tao},
  booktitle = {ICCV},
  year      = {2025},
  note      = {arXiv:2406.14540}
}

@article{li2025uva,
  title   = {Unified Video Action Model},
  author  = {Li, Shuang and Gao, Yihuai and Sadigh, Dorsa and Song, Shuran},
  journal = {arXiv preprint arXiv:2503.00200},
  year    = {2025},
  url     = {https://arxiv.org/abs/2503.00200}
}

@inproceedings{kim2026cosmos,
  title     = {Cosmos Policy: Fine-Tuning Video Models for Visuomotor Control and Planning},
  author    = {Moo Jin Kim and Yihuai Gao and Tsung-Yi Lin and Yen-Chen Lin and Yunhao Ge and Grace Lam and Percy Liang and Shuran Song and Ming-Yu Liu and Chelsea Finn and Jinwei Gu},
  booktitle = {The Fourteenth International Conference on Learning Representations},
  year      = {2026},
  url       = {https://openreview.net/forum?id=wPEIStHxYH}
}

@misc{ye2026worldactionmodelszeroshot,
  title         = {World Action Models are Zero-shot Policies},
  author        = {Seonghyeon Ye and Yunhao Ge and Kaiyuan Zheng and Shenyuan Gao and Sihyun Yu and George Kurian and Suneel Indupuru and You Liang Tan and Chuning Zhu and Jiannan Xiang and Ayaan Malik and Kyungmin Lee and William Liang and Nadun Ranawaka and Jiasheng Gu and Yinzhen Xu and Guanzhi Wang and Fengyuan Hu and Avnish Narayan and Johan Bjorck and Jing Wang and Gwanghyun Kim and Dantong Niu and Ruijie Zheng and Yuqi Xie and Jimmy Wu and Qi Wang and Ryan Julian and Danfei Xu and Yilun Du and Yevgen Chebotar and Scott Reed and Jan Kautz and Yuke Zhu and Linxi {``Jim''} Fan and Joel Jang},
  year          = {2026},
  eprint        = {2602.15922},
  archivePrefix = {arXiv},
  primaryClass  = {cs.RO},
  url           = {https://arxiv.org/abs/2602.15922}
}

@article{worldmem,
  title   = {WorldMem: Long-term Consistent World Simulation with Memory},
  author  = {Xiao, Zeqi and Lan, Yushi and Zhou, Yifan and Ouyang, Wenqi and Yang, Shuai and Zeng, Yanhong and Pan, Xingang},
  journal = {arXiv preprint arXiv:2504.12369},
  year    = {2025}
}

@article{zhu2025uwm,
  title   = {Unified World Models: Coupling Video and Action Diffusion for Pretraining on Large Robotic Datasets},
  author  = {Zhu, Chuning and Yu, Raymond and Feng, Siyuan and Burchfiel, Benjamin and Shah, Paarth and Gupta, Abhishek},
  journal = {arXiv preprint arXiv:2504.02792},
  year    = {2025},
  url     = {https://arxiv.org/abs/2504.02792}
}

@article{li2026dworldeval,
  title   = {dWorldEval: Scalable Robotic Policy Evaluation via Discrete Diffusion World Model},
  author  = {Li, Yaxuan and Zhou, Zhongyi and Chen, Yefei and Xue, Yaokai and Zhu, Yichen},
  journal = {arXiv preprint arXiv:2604.22152},
  year    = {2026},
  url     = {https://arxiv.org/abs/2604.22152}
}

@article{wang2026interactive,
  title   = {Interactive World Simulator for Robot Policy Training and Evaluation},
  author  = {Wang, Yixuan and Syed, Rhythm and Wu, Fangyu and Zhang, Mengchao and Onol, Aykut and Barreiros, Jose and Nayyeri, Hooshang and Dear, Tony and Zhang, Huan and Li, Yunzhu},
  journal = {arXiv preprint arXiv:2603.08546},
  year    = {2026},
  url     = {https://arxiv.org/abs/2603.08546}
}

@article{pi05,
  title   = {{$\pi_{0.5}$}: A Vision-Language-Action Model with Open-World Generalization},
  author  = {{Physical Intelligence} and Black, Kevin and Brown, Noah and Darpinian, James and Dhabalia, Karan and Driess, Danny and Esmail, Adnan and Equi, Michael and Finn, Chelsea and Fusai, Niccolo and Galliker, Manuel Y. and Ghosh, Dibya and Groom, Lachy and Hausman, Karol and Ichter, Brian and Jakubczak, Szymon and Jones, Tim and Ke, Liyiming and LeBlanc, Devin and Levine, Sergey and Li-Bell, Adrian and Mothukuri, Mohith and Nair, Suraj and Pertsch, Karl and Ren, Allen Z. and Shi, Lucy Xiaoyang and Smith, Laura and Springenberg, Jost Tobias and Stachowicz, Kyle and Tanner, James and Vuong, Quan and Walke, Homer and Walling, Anna and Wang, Haohuan and Yu, Lili and Zhilinsky, Ury},
  journal = {arXiv preprint arXiv:2504.16054},
  year    = {2025},
  url     = {https://arxiv.org/abs/2504.16054}
}

@inproceedings{morel,
  title     = {{MOReL}: Model-Based Offline Reinforcement Learning},
  author    = {Kidambi, Rahul and Rajeswaran, Aravind and Netrapalli, Praneeth and Joachims, Thorsten},
  booktitle = {Advances in Neural Information Processing Systems (NeurIPS)},
  year      = {2020},
  url       = {https://arxiv.org/abs/2005.05951}
}

@inproceedings{mopo,
  title     = {{MOPO}: Model-Based Offline Policy Optimization},
  author    = {Yu, Tianhe and Thomas, Garrett and Yu, Lantao and Ermon, Stefano and Zou, James and Levine, Sergey and Finn, Chelsea and Ma, Tengyu},
  booktitle = {Advances in Neural Information Processing Systems (NeurIPS)},
  year      = {2020},
  url       = {https://arxiv.org/abs/2005.13239}
}

@inproceedings{pets,
  title     = {Deep Reinforcement Learning in a Handful of Trials Using Probabilistic Dynamics Models},
  author    = {Chua, Kurtland and Calandra, Roberto and McAllister, Rowan and Levine, Sergey},
  booktitle = {Advances in Neural Information Processing Systems (NeurIPS)},
  year      = {2018},
  url       = {https://arxiv.org/abs/1805.12114}
}

@inproceedings{lipman2023flow,
  title     = {Flow Matching for Generative Modeling},
  author    = {Lipman, Yaron and Chen, Ricky T. Q. and Ben-Hamu, Heli and Nickel, Maximilian and Le, Matt},
  booktitle = {International Conference on Learning Representations (ICLR)},
  year      = {2023},
  url       = {https://arxiv.org/abs/2210.02747}
}

@inproceedings{liu2023flow,
  title     = {Flow Straight and Fast: Learning to Generate and Transfer Data with Rectified Flow},
  author    = {Liu, Xingchao and Gong, Chengyue and Liu, Qiang},
  booktitle = {International Conference on Learning Representations (ICLR)},
  year      = {2023},
  url       = {https://arxiv.org/abs/2209.03003}
}

@article{wmuncertainty25,
  title   = {World Models That Know When They Don't Know: Controllable Video Generation with Calibrated Uncertainty},
  author  = {Mei, Zhiting and Yin, Tenny and Baker, Micah and Shorinwa, Ola and Majumdar, Anirudha},
  journal = {arXiv preprint arXiv:2512.05927},
  year    = {2025},
  url     = {https://arxiv.org/abs/2512.05927}
}

@article{wav26,
  title   = {World Action Verifier: Self-Improving World Models via Forward-Inverse Asymmetry},
  author  = {Liu, Yuejiang and Feng, Fan and Kong, Lingjing and Lu, Weifeng and Tang, Jinzhou and Zhang, Kun and Murphy, Kevin and Finn, Chelsea and Du, Yilun},
  journal = {arXiv preprint arXiv:2604.01985},
  year    = {2026},
  url     = {https://arxiv.org/abs/2604.01985}
}

@article{cosmos3,
  title   = {Cosmos 3: Omnimodal World Models for Physical AI},
  author  = {{NVIDIA} and others},
  journal = {arXiv preprint arXiv:2606.02800},
  year    = {2026},
  url     = {https://arxiv.org/abs/2606.02800}
}
